\documentclass[]{typhoon}

\usepackage{amsmath,amssymb}
\usepackage{float}
\usepackage{enumitem}
\usepackage{tabularx}
\usepackage{xspace}
\usetikzlibrary{positioning,arrows.meta,fit,calc}

\newcommand{\huggingface}{\raisebox{-1.5pt}{\includegraphics[height=1.05em]{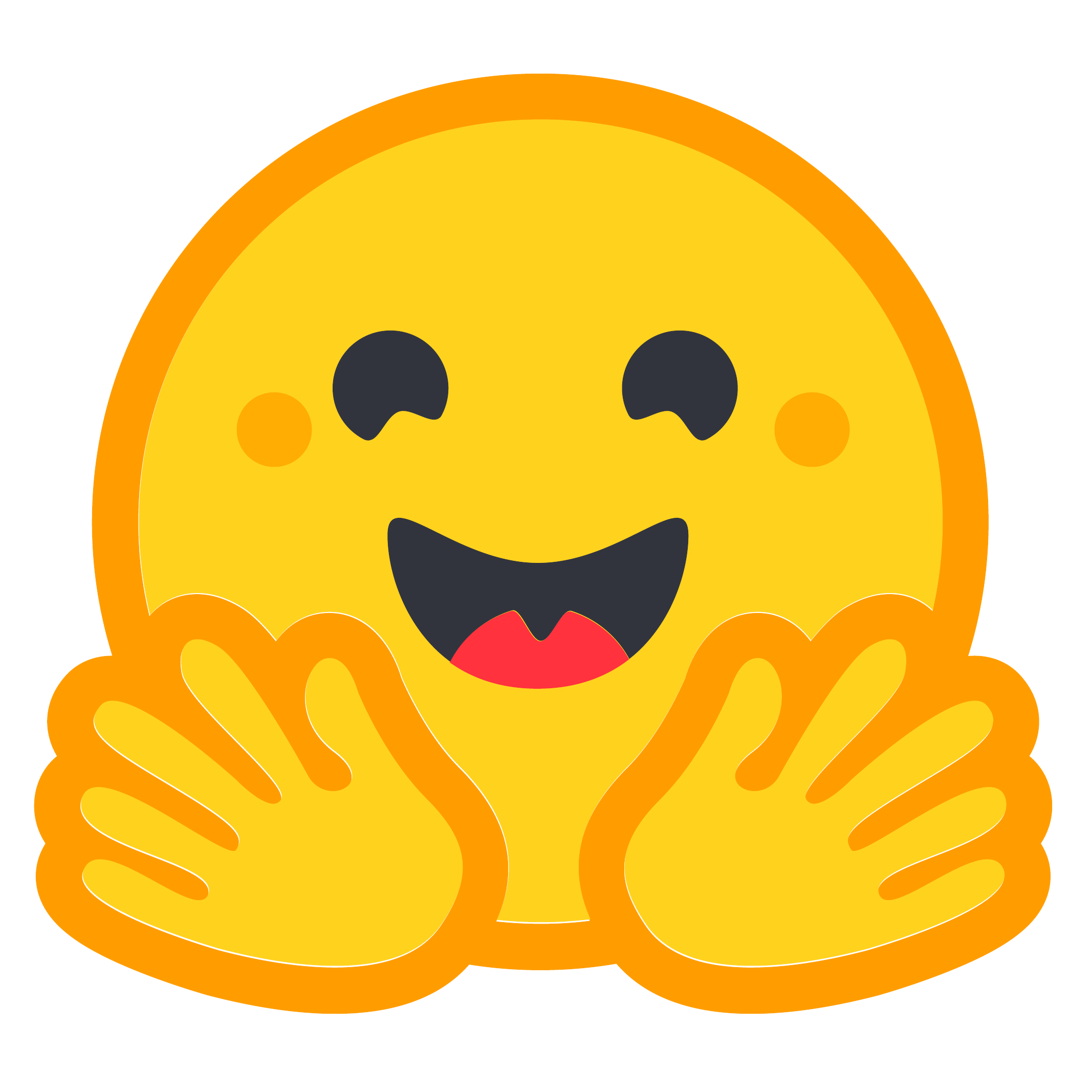}}\xspace}
\newcommand{\github}{\raisebox{-1.5pt}{\includegraphics[height=1.05em]{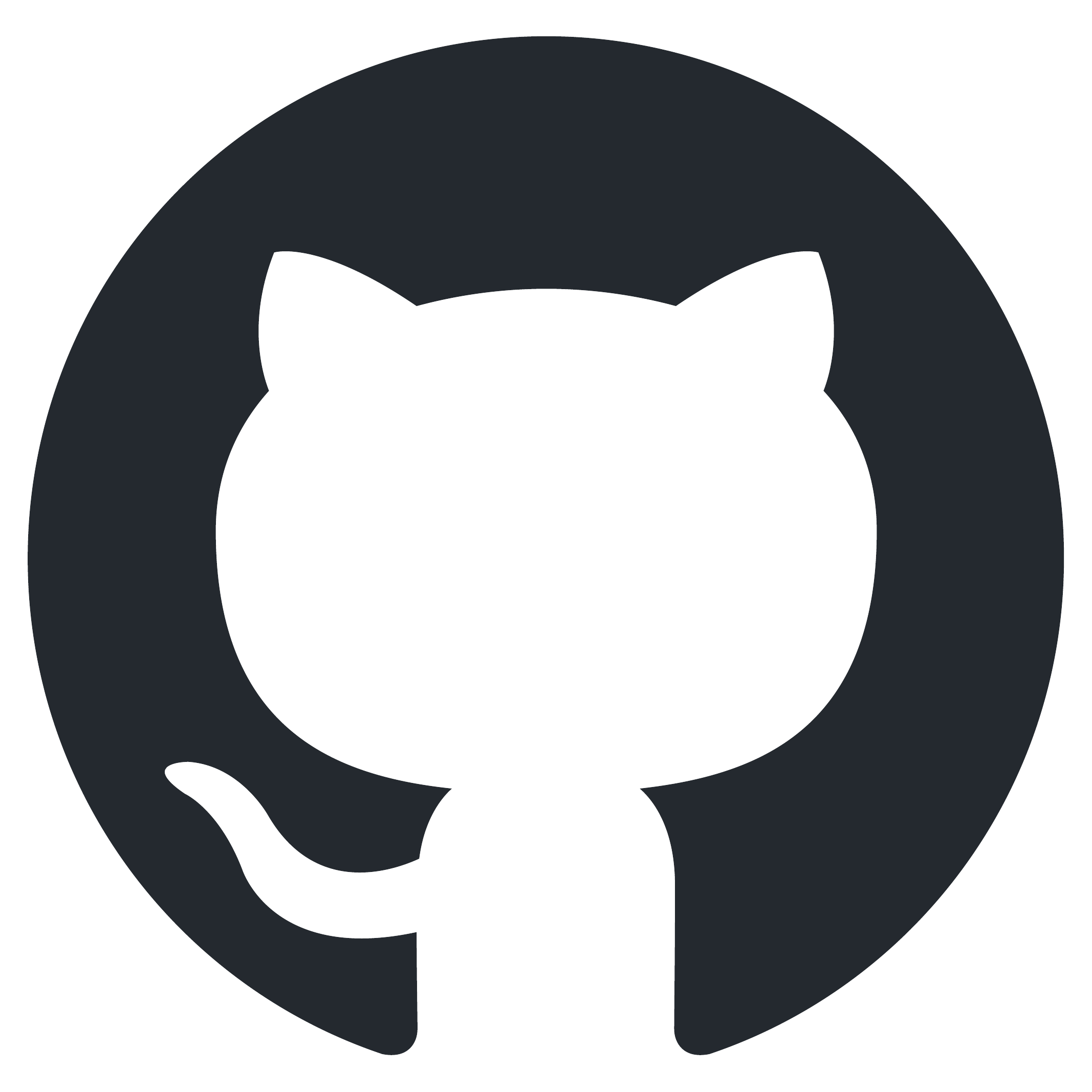}}\xspace}

\newcommand{\tht}[1]{\textthai{#1}}
\newcommand{\khammoon}{khammoon\xspace}

\setlist[itemize]{leftmargin=*}
\setlist[enumerate]{leftmargin=*}

\title{khammoon: From a Linguistic Guideline to a Thai Morpheme Segmenter}

\author[1]{Warit Sirichotedumrong}
\author[1]{Adisai Na-Thalang}
\author[1]{Chanakan Wittayasakpan}
\author[1]{Kritsadha Phatcharoen}
\author[2]{Kunat Pipatanakul}
\affiliation[1]{Typhoon, SCB DataX}
\affiliation[2]{AI Singapore}
\abstract{
Thai is written without spaces between words, making segmentation a foundational step in most Thai NLP pipelines. Existing tools primarily reproduce the word-level conventions of corpora such as BEST; however, because the notion of a ``word'' in Thai lacks a unified linguistic consensus, these conventions struggle with productive compounding, leading to arbitrary boundaries and ambiguity in downstream representations. Instead, we propose operating at the morphological level, focusing on  \emph{khammoon} or \tht{คำมูล}, the smallest atomic unit that carries meaning. Segmenting at the morphemic level provides an explicit, consistent morphological representation that eliminates compounding ambiguity. Crucially, this representation benefits downstream learning: models trained on morphemes perform on par with or exceed word-level baselines across part-of-speech tagging, named-entity recognition, and sentiment classification, demonstrating that linguistic consistency improves representation quality without downstream performance penalty. To operationalize this representation, we translate linguistic principles into a formal decision guideline with near-perfect human agreement ($\kappa \approx 0.99$), scale annotation with an LLM, and correct systematic errors with human labels. Finally, we train \khammoon, a lightweight 1M-parameter convolutional BiLSTM segmenter designed for efficient CPU inference, achieving a boundary F1 of 0.992 on out-of-domain text. The guideline, datasets, model, and library are all publicly released.
}

\checkdata[Project Resources]{\newline
\github\texttt{Code:} \url{https://github.com/warit-s/khammoon-segmenter} \newline
\huggingface\texttt{Model:} \url{https://huggingface.co/typhoon-ai/khammoon} \newline
\huggingface\texttt{Data:} \url{https://huggingface.co/datasets/typhoon-ai/khammoon-data}
}

\begin{document}
\maketitle
\vspace{-20pt}

\section{Introduction}
\label{sec:intro}
Thai is written without spaces between words, making word segmentation a fundamental step in Thai natural language processing (NLP). Existing tools, including the dictionary-based newmm in PyThaiNLP \citep{phatthiyaphaibun2023pythainlp} and neural models such as DeepCut \citep{deepcut2019} and AttaCut \citep{chormai2019attacut}, primarily operate at the word level, following conventions established by resources such as the BEST corpus \citep{kosawat2009best}. While word segmentation is useful for many downstream tasks, word boundaries do not necessarily reflect the internal morphological structure of Thai, thereby introducing ambiguity in how compounds are represented. In particular, Thai allows productive compounding, and the distinction between words and compounds is not always clear-cut \citep{aroonmanakun2007thoughts}. A finer-grained representation based on Thai morphemes, or khammoon (\tht{คำมูล}), can make this structure explicit by identifying the smallest meaningful units. For example, \tht{สินค้า} `goods' consists of \tht{สิน} `wealth' and \tht{ค้า} `trade', whereas \tht{ราคา} `price' is not further divisible into meaningful components. Segmenting at this level offers a consistent linguistic basis for representing morphological composition, complementing existing word-level tools and enabling applications that benefit from access to subword meaning. \Cref{fig:teaser} illustrates the distinction between word-level and morpheme segmentation.

Our contributions are as follows:
\begin{enumerate}[nosep]
  \item \textbf{An operational definition of the Thai morpheme.} We develop a linguistically grounded annotation guideline that translates the concept
    of a Thai morpheme into explicit tests and decision rules. Three trained annotators
    apply the guideline to 4,252 sentences from two domains, achieving near-perfect inter-annotator agreement (Fleiss' $\kappa \approx 0.99$).
  \item \textbf{An efficient Thai morpheme segmenter.} We use the guideline and human-labelled examples to scale annotation to a 70M-character corpus with an LLM, correct systematic errors in the generated labels, and train \khammoon, a 1M-parameter convolutional BiLSTM character tagger that runs on a CPU.

  \item \textbf{Public releases:} We release the linguistic guideline, human-labelled development and test sets,
the LLM-annotated corpus with original and corrected labels, the trained model,
and the library to support further research and development.
\end{enumerate}

\begin{figure}[!tb]
  \centering
  \small
  \renewcommand{\arraystretch}{1.25}
  \begin{tabular}{@{}l l@{}}
\toprule
Reference & \tht{กรรมวิธี\textcolor{typhoonpurple}{\,|\,}ที่\textcolor{typhoonpurple}{\,|\,}ใช้\textcolor{typhoonpurple}{\,|\,}แยก\textcolor{typhoonpurple}{\,|\,}ก๊าซ\textcolor{typhoonpurple}{\,|\,}ออก\textcolor{typhoonpurple}{\,|\,}จาก\textcolor{typhoonpurple}{\,|\,}น้ำ\textcolor{typhoonpurple}{\,|\,}มัน\textcolor{typhoonpurple}{\,|\,}ดิบ} \\
\midrule
PyThaiNLP newmm & \tht{กรรมวิธี\textcolor{typhoonpurple}{\,|\,}ที่\textcolor{typhoonpurple}{\,|\,}ใช้\textcolor{typhoonpurple}{\,|\,}แยก\textcolor{typhoonpurple}{\,|\,}ก๊าซ\textcolor{typhoonpurple}{\,|\,}ออกจาก\textcolor{typhoonpurple}{\,|\,}น้ำมันดิบ} \\
AttaCut & \tht{กรรมวิธี\textcolor{typhoonpurple}{\,|\,}ที่\textcolor{typhoonpurple}{\,|\,}ใช้\textcolor{typhoonpurple}{\,|\,}แยก\textcolor{typhoonpurple}{\,|\,}ก๊าซ\textcolor{typhoonpurple}{\,|\,}ออก\textcolor{typhoonpurple}{\,|\,}จาก\textcolor{typhoonpurple}{\,|\,}น้ำมัน\textcolor{typhoonpurple}{\,|\,}ดิบ} \\
LLM segmenter & \tht{กรรม\textcolor{typhoonpurple}{\,|\,}วิธี\textcolor{typhoonpurple}{\,|\,}ที่\textcolor{typhoonpurple}{\,|\,}ใช้\textcolor{typhoonpurple}{\,|\,}แยก\textcolor{typhoonpurple}{\,|\,}ก๊าซ\textcolor{typhoonpurple}{\,|\,}ออก\textcolor{typhoonpurple}{\,|\,}จาก\textcolor{typhoonpurple}{\,|\,}น้ำ\textcolor{typhoonpurple}{\,|\,}มัน\textcolor{typhoonpurple}{\,|\,}ดิบ} \\
\khammoon & \tht{กรรมวิธี\textcolor{typhoonpurple}{\,|\,}ที่\textcolor{typhoonpurple}{\,|\,}ใช้\textcolor{typhoonpurple}{\,|\,}แยก\textcolor{typhoonpurple}{\,|\,}ก๊าซ\textcolor{typhoonpurple}{\,|\,}ออก\textcolor{typhoonpurple}{\,|\,}จาก\textcolor{typhoonpurple}{\,|\,}น้ำ\textcolor{typhoonpurple}{\,|\,}มัน\textcolor{typhoonpurple}{\,|\,}ดิบ} \\
\bottomrule
\end{tabular}
  \caption{\textbf{One sentence, four segmenters.} A sentence from the BEST2009 development set, ``the process used
  to separate gas from crude oil''. The word segmenters keep the compounds \tht{ออกจาก} `out of' and
  \tht{น้ำมัน} `oil', from \tht{น้ำ} `water' and \tht{มัน} `fat', whole; the LLM segmenter splits
  \tht{กรรมวิธี} `process', a Pali--Sanskrit compound the guideline keeps; \khammoon matches the
  reference.}
  \label{fig:teaser}
\end{figure}

\section{Related Work}
\label{sec:related}
\paragraph{Thai word segmentation.} The BEST corpus \citep{kosawat2009best} fixed the word conventions
that Thai segmenters have followed since. Dictionary-based segmenters such as newmm in PyThaiNLP
\citep{phatthiyaphaibun2023pythainlp} find the longest matching words; neural segmenters such as
DeepCut \citep{deepcut2019}, a convolutional character tagger, and AttaCut \citep{chormai2019attacut},
which adds syllable features for speed, learn the BEST conventions from data; SEFR and OSKut
\citep{limkonchotiwat2020sefr,limkonchotiwat2021oskut} transfer them to new domains with little
labelled data. \citet{aroonmanakun2007thoughts} discusses why the word itself is hard to define in Thai. While this work provides a linguistic perspective on defining word boundaries using the notion of true compounds, distinguishing true compounds in practice remains challenging and prone to inconsistencies, such as classifying \tht{แม่น้ำ} `river' from \tht{แม่} `creator' and \tht{น้ำ} `water' as a true compound while treating \tht{หม้อ|หุง|ข้าว} `rice cooker' from \tht{หม้อ} `pot', \tht{หุง} `cook' and \tht{ข้าว} `rice' separately. We argue that semantic transparency is inherently a gradient, making it difficult to establish a definitive threshold for how `opaque' a compound must be to qualify as a true compound. All of these systems take the word as their unit, and whether a string counts as one word
depends on the conventions the BEST annotators agreed on, which differ from compound to compound.
\khammoon keeps the character-tagging design of DeepCut but changes the unit to the morpheme,
where the same tests from the guideline of \Cref{sec:guideline}, such as whether another word can
replace one of the parts, decide every boundary.

\paragraph{Units below the word.} Thai Character Clusters \citep{theeramunkong2000tcc} are the
smallest groups of characters that cannot be split without breaking the spelling, and syllable
segmenters split text one level above them; both are orthographic units without meaning, and we use
the former as a hard constraint. Subword tokenizers such as byte-pair encoding
\citep{sennrich-etal-2016-neural} and SentencePiece \citep{kudo-richardson-2018-sentencepiece}
learn units from frequency statistics, and Thai language models typically apply them to raw or
word-segmented text \citep{lowphansirikul2021wangchanberta}. The Thai morpheme sits between these
two: it is a meaning-bearing unit like a word, but closed and stable like a character inventory.

\paragraph{LLMs as annotators.} LLMs have been used to label training data in place of crowd
workers \citep{wang-etal-2021-want-reduce,ding-etal-2023-gpt,gilardi2023chatgpt}, usually with a
small model distilled from the labels \citep{hinton2015distilling}; majority voting over sampled
answers improves the reliability of such labels \citep{wang2023selfconsistency}. We follow this
recipe and add one step: in our setting the LLM fails systematically on one class of case rather
than at random, so we correct its labels with a small amount of human-labelled data before training
the small model.

\section{A Linguistic Guideline for Thai Morphemes}
\label{sec:guideline}
This section presents the linguistic rationale behind our guideline, explaining how Thai words are formed, why native speakers may disagree on word boundaries, and how these considerations inform the segmentation rules. The full guideline is in Appendix~\ref{app:guideline}.

\subsection{Thai word formation and why boundaries are disputed}
Thai builds new words by putting units together, not by inflection. Traditional Thai grammar
distinguishes the morpheme, \tht{คำมูล}, the smallest unit that can stand alone with a
complete meaning, such as \tht{บ้าน} `house', \tht{ราคา} `price' and \tht{นาฬิกา} `clock', from
several kinds of built words: compounds that keep the meaning of their parts, such as \tht{กรงนก}
`bird cage' from \tht{กรง} `cage' and \tht{นก} `bird'; compounds with a new meaning, such as
\tht{รถไฟ} `train' from \tht{รถ} `vehicle' and \tht{ไฟ} `fire'; couplets, \tht{คำซ้อน}, of
synonyms or opposites, such as \tht{ดีชั่ว} `good and evil'; reduplication, \tht{คำซ้ำ}, as in
\tht{ดำๆ} `blackish'; and Pali--Sanskrit compounds, \tht{คำสมาส} and \tht{คำสนธิ}, which are read
back to front, as in \tht{แพทยศาสตร์} `medical science' from \tht{แพทย} `physician' and
\tht{ศาสตร์} `science', and which may fuse sounds at the join, as \tht{วิทยาลัย} `college' does
from \tht{วิทย} and \tht{อาลัย}. Note that the term `morpheme' in this article refers specifically to the Thai concept of morpheme and may differ from the concept of morpheme in general linguistics (particularly, the general linguistic concept of `morpheme' also includes units with a complete meaning even if they cannot stand alone). Furthermore, classifying a word into its exact category is often difficult, and this category assignment directly influences the segmentation decision.

The rationale names the factors that make native speakers place boundaries differently in the same
text: how tightly the parts cohere, so that \tht{น้ำตก} `waterfall' feels like one word and \tht{น้ำ|ดื่ม}
`drinking water' like two; whether the meaning follows from the parts, which it does for \tht{มือซ้าย} `left
hand' but not for \tht{มือถือ} `mobile phone'; how many parts there are; whether a part is shared
between neighbouring compounds, so that \tht{รถไฟฟ้า} `electric train' can be read as
\tht{รถไฟ}+\tht{ฟ้า} or \tht{รถ}+\tht{ไฟฟ้า}; whether the speaker still recognizes a historical
part such as \tht{สอ} in \tht{ดินสอ} `pencil'; sound change that leaves a part meaningless, as with
\tht{ตะ} in \tht{ตะวัน} `sun', from \tht{ตา} `eye'; and context, which decides between \tht{ข้าวเย็น} `dinner' and
\tht{ข้าว|เย็น} `cold rice', or between \tht{รอบ|อก} `chest measurement' and \tht{รอ|บอก} `wait
to tell' in the same string of letters. Word segmentation standards settle such cases by
convention; the guideline settles them by definition.

\subsection{Definition, tests and decision procedure}
A string is a morpheme if no part of it can be replaced or separated without losing or changing
its meaning. Two tests make this operational. In the \emph{substitution test}, if another word can
take the place of one part, as \tht{มือปืน} `gunman' and \tht{นับถือ} `to respect' do for
\tht{มือ|ถือ}, the string is not a morpheme. In the \emph{insertion test}, if a word can be
inserted between the parts with the meaning preserved, as in \tht{กรงใส่นก} `cage for birds' for
\tht{กรง|นก}, the string is not a morpheme; if insertion destroys the meaning, as it does for
\tht{ทำลาย} `to destroy', which comes from \tht{ทลาย}, it is. When the tests are inconclusive the annotator consults a dictionary for the
etymology, and the meaning in context is fixed before the tests are applied: \tht{ทำลาย} is one
unit when it means `destroy' and two units, \tht{ทำ|ลาย} `make a pattern', otherwise.

The annotation guideline turns this into four steps, applied in order to every input, under one
hard constraint: the tokens must concatenate back to the input exactly, so annotators only insert
boundaries and never alter characters.
\begin{enumerate}[nosep]
  \item \textbf{Structural segmentation.} Split at phrase and clause boundaries; every whitespace
        character becomes its own token.
  \item \textbf{Non-Thai text, punctuation and symbols.} URLs, e-mail addresses, file names,
        numbers with separators, heading markers such as \tht{(๒)} and chemical formulas are kept
        whole. Foreign-script text is split only at whitespace, so \texttt{People's} and
        \texttt{COVID-19} stay whole. Punctuation, symbols and mathematical operators are separated
        from adjacent text.
  \item \textbf{Thai-character handling.} Segment by the intended spelling even when the text is
        misspelled; keep an abbreviation of one entity such as \tht{ผบ.ทบ.} whole but split
        stacked titles such as \tht{รศ.|ดร.|พญ.}; separate the repetition mark \tht{ๆ} and the
        ellipsis mark \tht{ฯ}, as in \tht{กรุง|เทพ|ฯ}; and read the whole sentence to fix the meaning before Step~4.
  \item \textbf{Morpheme or compound.} Assign each remaining chunk to the first category of
        \Cref{tab:guideline} that applies and follow its rule.
\end{enumerate}
The guideline contains 151 worked examples, each an input with its correct segmentation: 104
illustrate individual rules and 47 are full sentences listed after the rules.

\begin{table*}[!t]
\centering
\caption{\textbf{Step 4 of the guideline: the five categories and their rules.} The first category
that applies decides. Examples are taken from the guideline; \textbar{} marks a boundary.}
\label{tab:guideline}
\small
\begin{tabularx}{\textwidth}{@{}p{3.1cm} X >{\raggedright\arraybackslash}p{3.4cm} >{\raggedright\arraybackslash}p{4.2cm}@{}}
\toprule
\textbf{Category} & \textbf{Rule} & \textbf{Keep} & \textbf{Split} \\
\midrule
1. Mixed origin & Parts from different sources (Thai, Pali--Sanskrit, Khmer, foreign) are split. & ---
  & \tht{ผล|ไม้}, \tht{ราช|วัง}, \tht{พล|เมือง} \\
\addlinespace
2. All Pali--Sanskrit & Keep when the meaning is read back to front, a true \tht{สมาส} or
  \tht{สนธิ}; split when the parts are in Thai order or form a couplet.
  & \tht{ประวัติศาสตร์}, \tht{พุทธศาสนา} & \tht{ศาสนา|พุทธ}, \tht{ประชา|ชน} \\
\addlinespace
3. Loanwords & Keep when read as in the source language, with the head last; split when the
  parts follow Thai order, with the head first. & \tht{วิปครีม}, \tht{คอนกรีตบล็อก} & \tht{คุกกี้|ช็อกโกแลต} \\
\addlinespace
4. Native Thai & A: keep sound-reduced forms. B: split meaningless affix syllables, except the
  fused prefixes \tht{ระ-}, \tht{กระ-}, \tht{ประ-}. C: keep the base form of an onomatopoeia,
  split explicit doubling. D: otherwise apply the substitution and insertion tests.
  & \tht{ตะวัน}, \tht{กระโดด}, \tht{คิกคัก}, \tht{นาฬิกา} & \tht{ตา|เตอ}, \tht{คิก|คิก|คัก|คัก},
    \tht{รถ|ไฟ}, \tht{มือ|ถือ} \\
\addlinespace
5. Unknown or still uncertain & Pronunciation test: keep when a part changes its pronunciation in
  the combination, split when it does not. & \tht{อนึ่ง} & \tht{ขะมัก|เขม้น}, \tht{พระ|สงฆ์} \\
\bottomrule
\end{tabularx}
\end{table*}

\subsection{Human annotation and agreement}
\label{sec:annotation}
Three trained annotators labelled the two reference sets with the guideline. To measure how
reproducible the guideline is, 400 sentences from each corpus were first annotated independently by
all three. We score agreement on every position between two characters as a three-way decision,
boundary or not. Fleiss' kappa, which corrects for chance agreement, is 0.986 on BEST and 0.991 on
Wangchan, and the boundary F1 between annotator pairs averages 0.990 and 0.993. At the level of
whole sentences, all three annotators produced the identical segmentation for 60.5\% of the BEST
sentences and 45.5\% of the Wangchan documents, which average 200 and 830 characters respectively,
so a single differing boundary counts as a mismatch. Disagreements concentrate in four groups:
Pali--Sanskrit compounds whose parts also exist as Thai words, such as \tht{เหตุการณ์} `event',
\tht{ประสบการณ์} `experience' and \tht{วัตถุประสงค์} `objective'; everyday compounds near the
keep/split line, such as \tht{หัวเราะ} `laugh';
proper names and transliterated brand names such as \tht{สยามแม็คโคร} and
\tht{โรบินสันไลฟ์สไตล์}; and multi-part abbreviations such as \tht{ผบช.ภ.}. Appendix~\ref{app:iaa} gives the per-pair figures and the full procedure.

The three versions of the 400 sentences were reconciled into one, and the remaining sentences were
divided among the annotators. A review pass by the linguists then edited 1,293 of the 4,252
sentences. Finally, a corpus-wide consistency check listed every pair of adjacent tokens that also
occurs merged elsewhere in the reference data; pairs with a clear majority were rewritten to the
majority form after the linguist confirmed each one, and pairs whose two forms are both legitimate
such as \tht{ทำลาย}, \tht{วิชาการ} and heading numbers like \tht{2)}, were resolved sentence by
sentence.
Every final sentence concatenates back to its source text exactly.

\section{\khammoon}
\label{sec:method}
We develop \khammoon, a small model for Thai morpheme segmentation. Our approach uses an
LLM to annotate a large corpus following the linguistic guideline,
corrects systematic errors in these annotations using human-labelled
examples, and trains a lightweight character-level segmenter on the
corrected data. This section describes the model architecture (\Cref{sec:model}),
training-data construction (\Cref{sec:data}), and training and inference procedures (\Cref{sec:training}).

\subsection{Model Architecture}
\label{sec:model}
\begin{figure}[!tb]
\centering
\begin{tikzpicture}[
  node distance=4mm,
  layer/.style={draw=typhoonpurple, very thick, rounded corners=2pt, fill=white, align=center,
                font=\footnotesize, minimum height=9mm, text width=78mm, inner sep=3pt},
  conv/.style={layer, text width=23mm},
  rule/.style={draw=black!60, rounded corners=2pt, fill=black!4, align=center, font=\footnotesize,
               minimum height=9mm, text width=78mm, inner sep=3pt},
  io/.style={align=center, font=\small},
  arr/.style={-{Latex[length=2mm]}, thick, color=black!70},
  lab/.style={font=\scriptsize, color=black!60, fill=white, inner sep=1pt}]
\node[io] (in) {Input characters: \tht{สินค้า}\\[1pt]
  {\footnotesize Thai Character Clusters: \tht{สิ}\enspace\tht{น}\enspace\tht{ค้า}}};
\node[layer, above=of in] (emb) {\textbf{Character embedding}, 64 dimensions\\
  206 frequent characters, plus 6 shared embeddings by character class};
\node[conv, above=7mm of emb] (c5) {\textbf{Convolution}\\ width 5, 128 channels};
\node[conv, left=4.5mm of c5] (c3) {\textbf{Convolution}\\ width 3, 128 channels};
\node[conv, right=4.5mm of c5] (c7) {\textbf{Convolution}\\ width 7, 128 channels};
\node[layer, above=7mm of c5] (lstm) {\textbf{Two-layer BiLSTM}, 128 units per direction};
\node[layer, above=of lstm] (lin) {\textbf{Linear layer}: one boundary score before each character};
\node[rule, above=of lin] (dec) {\textbf{Decoding rules}: boundary forced at whitespace, forbidden inside a
  cluster, otherwise kept if the score is at least 0.5};
\node[io, above=of dec] (out) {Output tokens: \tht{สิน}\,\textbar\,\tht{ค้า}};
\draw[arr] (in) -- (emb);
\foreach \c in {c3,c5,c7} \draw[arr] (emb.north -| \c) -- (\c.south);
\foreach \c in {c3,c5,c7} \draw[arr] (\c.north) -- (lstm.south -| \c);
\node[lab] at ($(c5.north)!0.5!(lstm.south)$) {ReLU, concatenated: 384};
\draw[arr] (lstm) -- (lin);
\draw[arr] (lin) -- (dec);
\draw[arr] (dec) -- (out);
\end{tikzpicture}
\caption{\textbf{Architecture of \khammoon.} Purple boxes are learned layers, with 1.06M parameters in
total; the grey box holds fixed rules. The input \tht{สินค้า} `goods' has six characters in three
clusters, so the rules leave the model two positions to decide, and it places a boundary at one of
them.}
\label{fig:arch}
\end{figure}
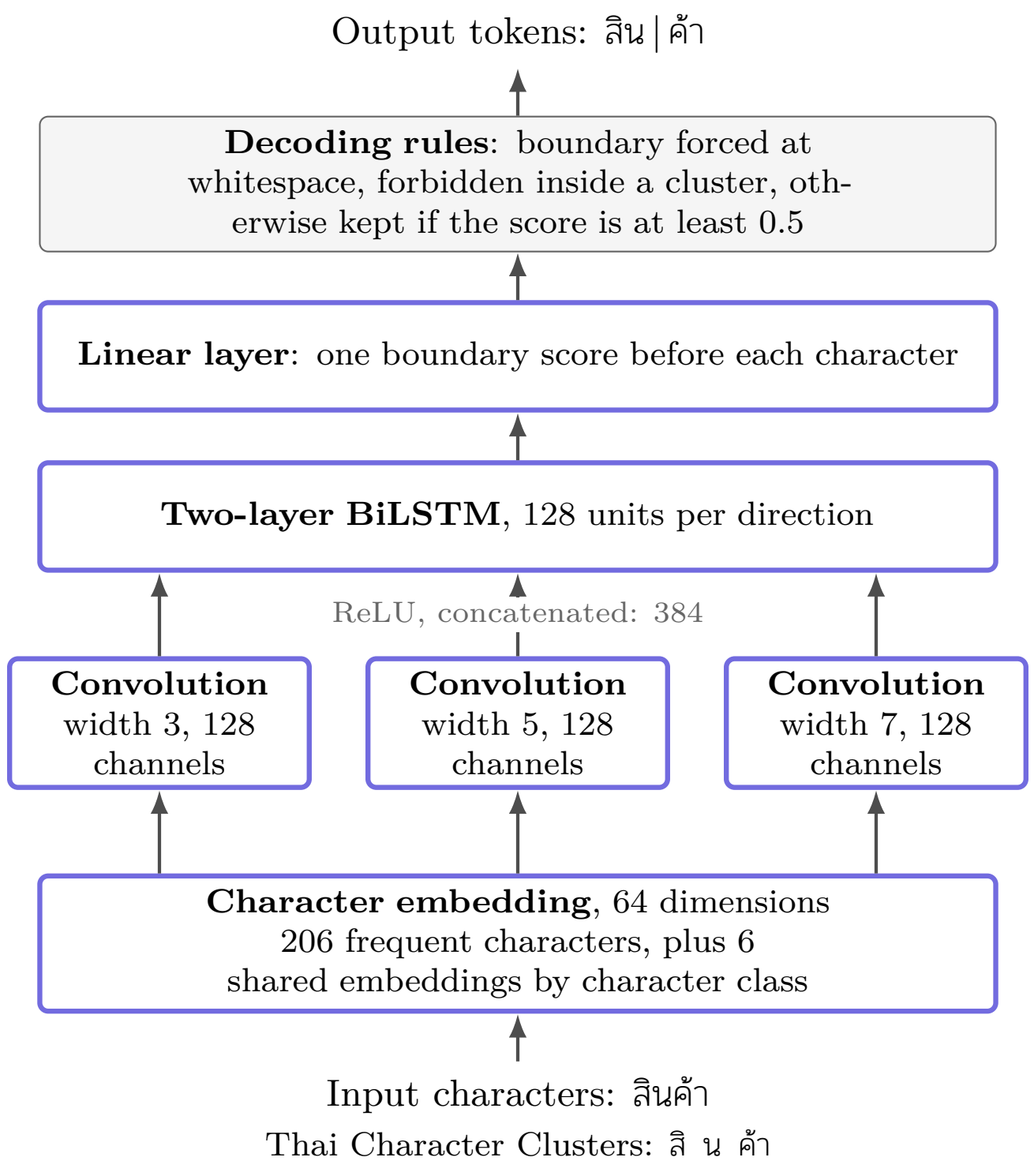

\khammoon is a convolutional BiLSTM model that treats Thai morpheme segmentation as a character-level binary classification task. The model predicts whether a morpheme boundary occurs before each character, subject to fixed orthographic constraints. As shown in \Cref{fig:arch}, boundaries are \emph{forced} at both edges of every whitespace character and \emph{forbidden} inside Thai Character Clusters (TCCs; \citealp{theeramunkong2000tcc}), which group characters that cannot be separated without breaking Thai spelling. For the remaining positions, the model embeds each character, applies three parallel convolutions of different widths, and passes their combined output through a two-layer BiLSTM. A linear layer then produces boundary scores. The model processes windows of 256 characters and is trained with binary cross-entropy loss at positions where boundaries are neither forced nor forbidden.

\subsection{Training data}
\label{sec:data}
We construct the training data in two steps. First, an LLM annotates the training corpus following the linguistic guideline (\Cref{sec:silver}). Second, we use the human-labelled development set to correct systematic errors in the LLM-generated annotations (\Cref{sec:relabel}). \Cref{tab:data} summarizes the dataset statistics.

\begin{table}[!tb]
\centering
\caption{\textbf{Dataset statistics.}}
\label{tab:data}
\begin{tabular}{l l r r r}
\toprule
\textbf{Part} & \textbf{Source} & \textbf{Rows} & \textbf{Characters} & \textbf{Boundaries} \\
\midrule
\rowcolor{typhoonpurple!20}\multicolumn{5}{l}{\textit{\textbf{LLM-labelled}}} \\
LLM-labelled & BEST2009 train & 147,999 & 20.9M & 4.89M \\
LLM-labelled & Wangchan train & 32,207 & 48.9M & 10.28M \\
\midrule
\rowcolor{typhoonpurple!20}\multicolumn{5}{l}{\textit{\textbf{Human-labelled}}} \\
development set (dev-train) & BEST2009 test & 1,484 & 305K & 70,766 \\
development set (dev-val) & BEST2009 test & 768 & 152K & 35,234 \\
test set & Wangchan test & 2,000 & 1.66M & 335,120 \\
\bottomrule
\end{tabular}
\end{table}

\subsubsection{Sources}
\label{sec:sources}
The two reference sets of \Cref{sec:guideline} come from different domains and play different
roles. 
The \textbf{development set} consists of 2,252 sentences from the
BEST2009 test split, covering news, novels, encyclopedia entries,
and articles. It is the only human-labelled data used during
development and is divided into two subsets: \emph{dev-train}
(1,484 sentences) is used to correct LLM-generated labels and
fine-tune the model, while \emph{dev-val} (768 sentences) is used
to select model checkpoints, mixing ratios, and the decision
threshold.
The \textbf{test set} contains 2,000 instruction--response documents
from the WangchanThaiInstruct test split
\citep{limkonchotiwat2025thaiinstruct}, covering finance, legal,
medical, and retail domains. It is reserved for final evaluation
to assess generalization to a different domain.

The \textbf{training corpus} combines the BEST2009 and
WangchanThaiInstruct training splits, totaling approximately
70M characters. To prevent data leakage, we remove 979 BEST2009
training rows that also appear in the reference sets, mostly
short repeated strings such as source URLs.

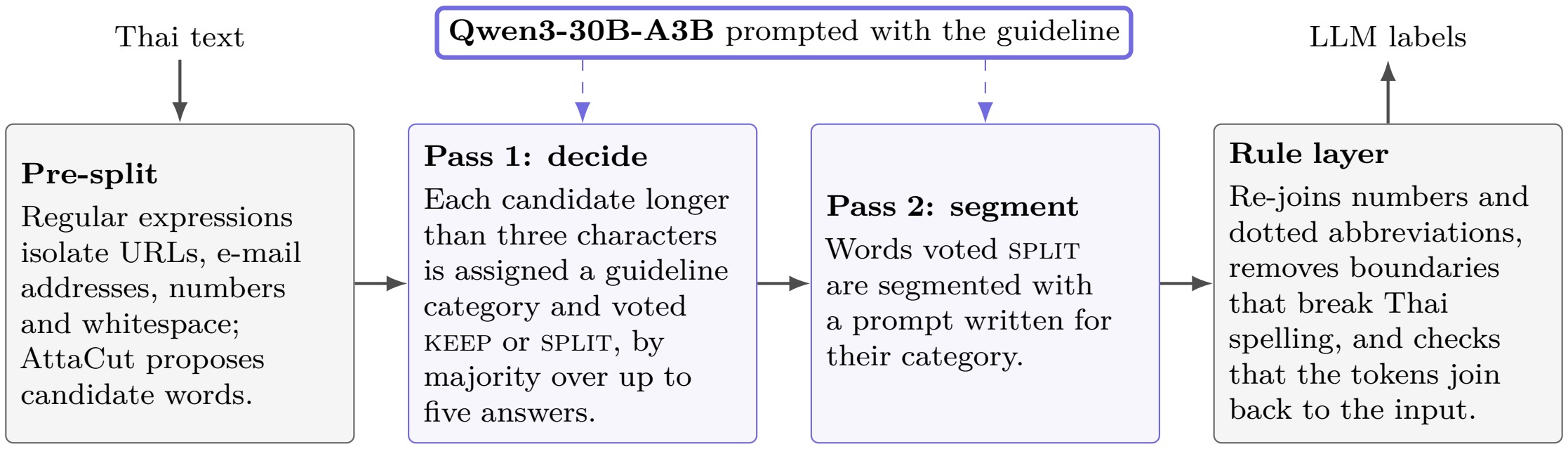
\begin{figure}[!tb]
\centering
\begin{tikzpicture}[
  node distance=4mm and 5mm,
  stagebox/.style={draw=typhoonpurple, rounded corners=2pt, align=left, fill=typhoonpurple!6,
               font=\footnotesize, execute at begin node={\hyphenpenalty=10000\relax},
               text width=30mm, minimum height=30mm, inner sep=4pt},
  rulebox/.style={stagebox, draw=black!60, fill=black!4},
  llm/.style={draw=typhoonpurple, very thick, rounded corners=2pt, fill=white, align=center,
              font=\footnotesize, inner sep=3pt},
  io/.style={align=center, font=\footnotesize},
  arr/.style={-{Latex[length=2mm]}, thick, color=black!70},
  larr/.style={-{Latex[length=2mm]}, color=typhoonpurple, dashed}]
\node[rulebox] (pre) {\textbf{Pre-split}\\[2pt] Regular expressions isolate URLs, e-mail addresses,
  numbers and whitespace; AttaCut proposes candidate words.};
\node[stagebox, right=of pre] (p1) {\textbf{Pass 1: decide}\\[2pt] Each candidate longer than three
  characters is assigned a guideline category and voted \textsc{keep} or \textsc{split}, by majority
  over up to five answers.};
\node[stagebox, right=of p1] (p2) {\textbf{Pass 2: segment}\\[2pt] Words voted \textsc{split} are
  segmented with a prompt written for their category.};
\node[rulebox, right=of p2] (post) {\textbf{Rule layer}\\[2pt] Re-joins numbers and dotted
  abbreviations, removes boundaries that break Thai spelling, and checks that the tokens join back
  to the input.};
\node[llm, above=6mm of $(p1.north)!0.5!(p2.north)$, minimum width=65mm]
  (llm) {\textbf{Qwen3-30B-A3B} prompted with the guideline};
\node[io, above=6mm of pre] (in) {Thai text};
\node[io, above=6mm of post] (out) {LLM labels};
\draw[arr] (in) -- (pre);
\draw[arr] (pre) -- (p1);
\draw[arr] (p1) -- (p2);
\draw[arr] (p2) -- (post);
\draw[arr] (post) -- (out);
\draw[larr] (llm.south -| p1) -- (p1.north);
\draw[larr] (llm.south -| p2) -- (p2.north);
\end{tikzpicture}
\caption{\textbf{The LLM segmenter that produced the training labels.} Grey boxes are fixed rules
and purple boxes are calls to the LLM, which receives the guideline in its prompt. Answers in
pass~1 are sampled at temperature 0.2.}
\label{fig:llm}
\end{figure}

\subsubsection{Labelling with the LLM}
\label{sec:silver}
To scale annotation to 70M characters, we use
Qwen3-30B-A3B-Instruct-2507 \citep{qwen3} to apply the
linguistic guideline to the training corpus. The annotation
pipeline consists of four steps, as shown in \Cref{fig:llm},
with majority voting in the first LLM pass following
\citet{wang2023selfconsistency}. The LLM-generated annotations achieve a boundary precision
of 0.977 and recall of 0.969 on the development set.
However, the errors show a systematic pattern: the LLM
often keeps common Thai compounds as single units when
the guideline requires them to be split. On the development
set, it misses 2,859 Thai-internal boundaries across 505
distinct strings, with the 20 most frequent strings
accounting for 39\% of these errors. Examples include
\tht{ชุม|ชน} `community', \tht{กฎ|หมาย} `law',
\tht{กรุง|เทพ|ฯ} `Bangkok', \tht{หนัง|สือ} `book',
\tht{ประเทศ|ไทย} `Thailand', and \tht{โครง|การ}
`project'. These recurring errors motivate the correction
procedure described in \Cref{sec:relabel}.

\begin{figure}[!tb]
\centering
\begin{tikzpicture}[
  node distance=4mm and 5mm,
  stage/.style={draw=typhoonpurple, rounded corners=2pt, align=left,
                font=\footnotesize, execute at begin node={\hyphenpenalty=10000\relax},
                text width=34mm, minimum height=33mm, inner sep=4pt, fill=typhoonpurple!6},
  model/.style={stage, fill=white, very thick},
  human/.style={draw=black!60, rounded corners=2pt, fill=black!4, align=center, font=\footnotesize,
                text width=40mm, inner sep=3pt},
  arr/.style={-{Latex[length=2mm]}, thick, color=black!70},
  harr/.style={-{Latex[length=2mm]}, color=black!50, dashed}]
\node[stage] (llm) {\textbf{Step 1: LLM labels}\\[2pt]
  70M characters labelled by the LLM segmenter with the guideline.\\[3pt]
  \tht{ชุมชน} kept whole 1,965 times\\
  \texttt{COVID-19} split all 1,115 times};
\node[stage, right=of llm] (r1) {\textbf{Step 2, round 1: copy the human conventions}\\[2pt]
  20,454 strings segmented consistently in dev-train are rewritten everywhere; Latin words split by
  the LLM are re-joined.\\[3pt]
  \tht{ชุม|ชน}, \texttt{COVID-19}};
\node[stage, right=of r1] (r2) {\textbf{Step 2, round 2: spread them}\\[2pt]
  A model fine-tuned on dev-train re-segments the corpus; only its boundaries between two Thai
  letters are kept.\\[3pt]
  Reaches strings that never occur in dev-train};
\node[model, right=of r2] (tr) {\textbf{Step 3: training \khammoon}\\[2pt]
  A new model learns from the round-2 labels, then is fine-tuned on batches that are half
  dev-train and half round-2 labels.};
\node[human, below=7mm of r2] (dev) {\textbf{dev-train}: 1,484 human-labelled sentences};
\draw[arr] (llm) -- (r1);
\draw[arr] (r1) -- (r2);
\draw[arr] (r2) -- (tr);
\draw[harr] (dev.west) -| (r1.south);
\draw[harr] (dev.north) -- (r2.south);
\draw[harr] (dev.east) -| (tr.south);
\end{tikzpicture}
\caption{\textbf{From LLM labels to the training data of \khammoon.} The LLM labels are corrected
twice with the human-labelled dev-train, marked by dashed arrows, before \khammoon is trained. The
counts in the first box are occurrences in the training text; after round~1, \tht{ชุมชน}
`community' is never kept whole and \texttt{COVID-19} is one token in 1,088 of its 1,115
occurrences.}
\label{fig:pipeline}
\end{figure}
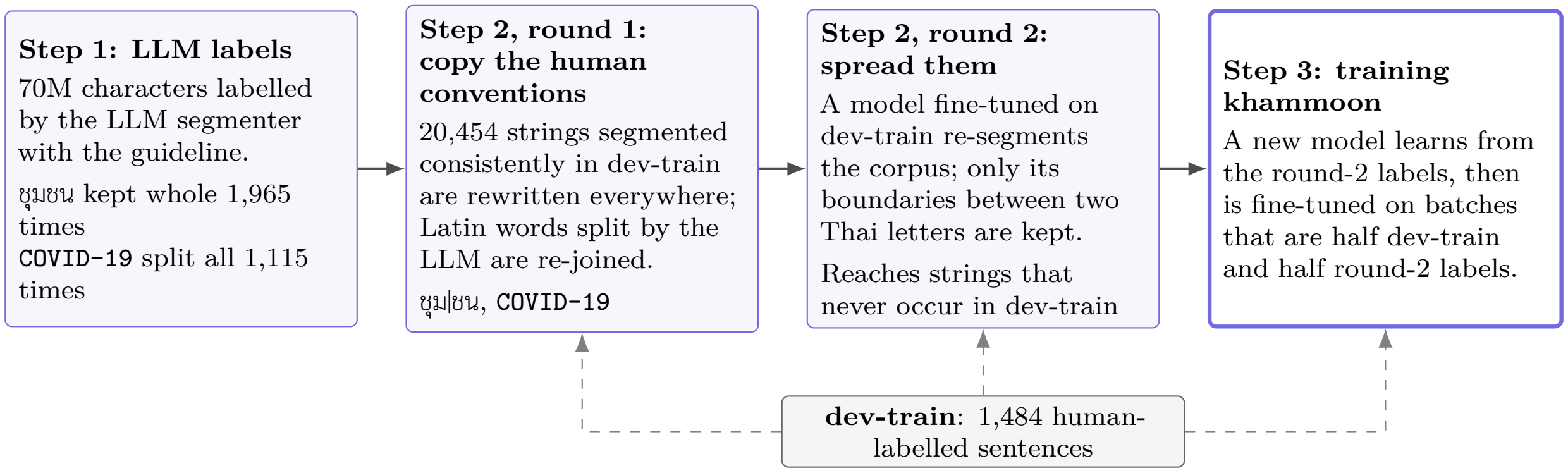

\subsubsection{Correcting the labels with human data}
\label{sec:relabel}

We correct systematic errors in the LLM-generated annotations
using the human-labelled dev-train set in two rounds, as shown
in \Cref{fig:pipeline}.

\textbf{Round 1: Direct correction.}
We extract Thai strings spanning one to four reference tokens
that occur at least twice in dev-train and have consistent
segmentations across all occurrences. We then replace their
segmentations in the LLM-labelled corpus with the human-labelled
versions. Following Step~2 of the guideline, we also rejoin
Latin words incorrectly split at hyphens or apostrophes,
such as \texttt{COVID-19} and \texttt{People's}, and
parenthesised markers such as \tht{(๒)}.

\textbf{Round 2: Model-based correction.}
To extend corrections to strings not found in dev-train,
we train a model with the architecture described in
\Cref{sec:model} on the round-1 labels and fine-tune it
on dev-train. We then use this model to re-segment the
training corpus, replacing the round-1 boundaries only
at positions between two Thai letters. This restriction
avoids changing non-Thai segmentation, for which the
human-labelled data provides limited coverage.

Round~1 adds 558,217 boundaries and removes 115,946,
while round~2 adds 101,130 and removes 130,006.

\subsection{Training and inference}
\label{sec:training}
\paragraph{Training.} We train \khammoon in two stages. First, we train the model on
the large-scale LLM-labelled corpus after two rounds of correction.
We then fine-tune it using a mixture of corrected LLM labels and
dev-train, a set of 1,484 human-labelled sentences, to better
align its predictions with the linguistic guideline. We found that
fine-tuning on human-labelled data alone reduces performance on
patterns absent from dev-train. For example, dev-train contains
no occurrences of \texttt{COVID-19} and consistently splits
hyphens, causing the model to segment it as \texttt{COVID|-|19}.
To address this, we draw half of each fine-tuning batch from
dev-train and half from the corrected LLM-labelled corpus.

We use dev-val, a separate set of 768 human-labelled sentences,
to select the mixing ratio, number of fine-tuning steps, and
decision threshold. We compare mixing ratios of 0\% and 50\%
LLM-labelled data and select the largest share whose boundary
F1 is within 0.0005 of the best result. We apply the same
selection criterion to the fine-tuning steps and decision
threshold, selecting a threshold of 0.5 from the range
0.30--0.70. \Cref{sec:ablations} reports these comparisons.

For the released model, we fine-tune on the full human-labelled
development set (dev-train and dev-val), mixed with the corrected
LLM-labelled corpus, and retain the weights from the final step.
\Cref{tab:hyperparameters} summarizes the training settings. Each run takes only a few minutes on a single H100 GPU. 

\begin{table}[!tb]
\centering
\caption{\textbf{Training settings.}}
\label{tab:hyperparameters}
\resizebox{\linewidth}{!}{%
\begin{tabular}{l c c}
\toprule
 & \textbf{Training} & \textbf{Fine-tuning (release)} \\
\midrule
Data & LLM labels & LLM-labelled + Human-labelled dev set\\
Steps & 6,000 & 1,000 \\
Batch (windows of 256 characters) & 256 & 64 \\
Peak learning rate & $2\times10^{-3}$ & $3\times10^{-4}$ \\
Warmup steps & 1,000 & 100 \\
Optimizer & \multicolumn{2}{c}{AdamW, $\beta=(0.9, 0.98)$, weight decay 0.01, cosine learning-rate schedule, bfloat16 precision} \\
\bottomrule
\end{tabular}}
\end{table}

\paragraph{Inference.} Since \khammoon processes at most 256 characters at a time,
longer texts are divided into overlapping windows, with 64
characters of context on each side. Each position uses the
prediction from the window where it is closest to the center,
reducing errors near window boundaries. To match the training
conditions, we add a newline at both ends of the input and
batch only windows of equal length, avoiding padding that
can affect predictions near the end of a text. We provide inference implementations using both ONNX Runtime and NumPy.

\section{Experimental Setup}
\label{sec:setup}
\paragraph{Metric.} We evaluate segmentation using boundary precision, recall,
and F1, computed over all character positions in each dataset.
A predicted boundary is correct if it matches the reference
boundary at the same position. We exclude boundaries adjacent
to whitespace because they are largely determined by the input
and inconsistently annotated. This makes our F1 scores lower
than evaluations that include whitespace boundaries.
We compute 95\% confidence intervals using 1,000 bootstrap
resamples of test documents, paired across systems.

\paragraph{Baselines.} We compare \khammoon against four groups of baselines:
\begin{itemize}[nosep]
  \item \textbf{Word segmenters:} newmm, nlpO3, DeepCut, and AttaCut.
  \item \textbf{Simple segmentation baselines:} the PyThaiNLP syllable segmenter and a dictionary-based segmenter that selects the most probable segmentation using a dictionary derived from LLM labels.
  \item \textbf{LLM segmenter:} the prompted LLM with our guideline.
  \item \textbf{Uncorrected BiLSTM:} the same architecture trained on the original LLM labels, without correction or fine-tuning.
\end{itemize}
Speed is measured as characters processed per second on one CPU thread on the 152K characters of
dev-val.

\paragraph{Downstream tasks.} Boundary F1 measures agreement with the linguistic guideline,
but not whether the resulting segments benefit downstream models.
We therefore evaluate each segmenter on two Thai downstream tasks
using the evaluation framework from our earlier tokenizer work.
Only the segmenter varies across configurations, and all models
are trained from scratch. For LST20, we align the original
word-level labels to the new segments using character offsets.
Each configuration is evaluated with ten random seeds.
\Cref{tab:downstream_setup} summarizes the tasks.

\begin{table}[!h]
\centering
\caption{\textbf{Downstream tasks.} Both models are trained from scratch on the output of
each segmenter.}
\label{tab:downstream_setup}
\small
\begin{tabularx}{\linewidth}{@{}l >{\raggedright\arraybackslash}X >{\raggedright\arraybackslash}X@{}}
\toprule
 & \textbf{LST20} \citep{boonkwan2020lst20} & \textbf{Wongnai} \\
\midrule
Task & POS tagging and NER & Five-class restaurant-review sentiment \\
Data & 63,310 / 5,620 / 5,250 training / validation / test sentences with word-level labels
     & 40,000 training reviews, 10\% held out for validation; official test set of 6,203 reviews \\
Model & Two-layer BiLSTM tagger: embedding 128, hidden 256, dropout 0.4
      & One-layer BiLSTM classifier: embedding 256, hidden 256, mean pooling, dropout 0.4, at
        most 256 tokens per review \\
Training & 30 epochs, early stopping at patience 5, batch 128, AdamW, learning rate $5\times10^{-4}$
         & 30 epochs, early stopping, batch 64, learning rate $3\times10^{-4}$ \\
Metric & Entity-level F1 for NER, token accuracy for POS & Accuracy and macro-F1 \\
\bottomrule
\end{tabularx}
\end{table}

\section{Results and Key Findings}
\label{sec:eval}
\subsection{Label correction and fine-tuning}
\label{sec:ablations}
\begin{table}[!h]
\centering
\caption{\textbf{Label correction and fine-tuning}, dev-val boundary F1, higher is better. $^\star$ marks the
configuration of the released model.}
\label{tab:ablation}
\begin{tabular}{l r r r}
\toprule
\textbf{LLM labels} & \textbf{No fine-tuning} & \textbf{Human-only fine-tuning} & \textbf{50\% LLM-labelled fine-tuning} \\
\midrule
uncorrected & 0.9772 & 0.9971 & 0.9928 \\
after round 1 & 0.9916 & 0.9970 & 0.9957 \\
after round 2 & 0.9955 & \textbf{0.9974} & 0.9972$^\star$ \\
\bottomrule
\end{tabular}
\end{table}
We evaluate the effects of label correction and fine-tuning
on dev-val, reserving the test set for the final evaluation
in \Cref{sec:mainresults}. \Cref{tab:ablation} compares
three versions of the LLM labels under three fine-tuning
settings. We highlight three findings.

\begin{itemize}[nosep]
  \item \textbf{Label correction improves performance.}
  Without fine-tuning, dev-val F1 increases from 0.9772
  with the original LLM labels to 0.9916 after round~1
  and 0.9955 after round~2.

  \item \textbf{Human-labelled data improves all models.}
  Fine-tuning only on dev-train raises F1 to
  0.9970--0.9974, regardless of the label version.
  However, this can cause the model to forget patterns
  absent from dev-train.

  \item \textbf{Corrected labels enable mixed fine-tuning.}
  Mixing LLM-labelled data with dev-train reduces F1
  relative to human-only fine-tuning by 0.0043 with
  uncorrected labels, 0.0013 with round-1 labels,
  and only 0.0002 with round-2 labels. Thus, label
  correction allows the model to retain broader
  training patterns with little loss on dev-val.
\end{itemize}

We also explored training without human-labelled data,
using guideline-derived rules and self-training.
These approaches were less effective: the best model
achieved 0.965 recall on Wangchan, while self-training
provided no further improvement. In particular,
corpus-based substitution rules could not reliably
distinguish meaningful compounds such as \tht{สิน|ค้า}
(`goods') from single morphemes such as \tht{อาหาร}
(`food') and \tht{ราคา} (`price'). These distinctions
often require etymological knowledge that cannot be
inferred from character co-occurrence alone.
In contrast, the final approach, which incorporates
human-labelled data, achieves 0.993 recall on Wangchan.

\subsection{Model size and the TCC constraint}
\label{sec:size_tcc}
\textbf{A larger model does not help.}
We compare three architectures trained on the same round-2
labels, with equal training steps and no fine-tuning.
The 1.06M-parameter BiLSTM achieves 0.9955 F1 on dev-val,
compared with 0.9957 for a 1.6M-parameter dilated CNN
and 0.9956 for a 25.6M-parameter Transformer.
Since larger models provide negligible improvement,
we retain the smaller BiLSTM.

\textbf{The TCC constraint removes a whole class of errors at no cost in recall.}
Removing the TCC constraint reduces dev-val precision
from 0.997 to 0.961 and F1 from 0.9972 to 0.979,
while recall remains nearly unchanged (0.9974 vs.\ 0.9976).
Without the constraint, the model produces 1,058 tokens
with invalid Thai character sequences, such as tokens
beginning with dependent vowels or tone marks.
The constraint excludes only 11 of 142,025 reference
boundaries in the development set, limiting the maximum
achievable recall by less than 0.01\%.
Thus, the TCC constraint substantially reduces invalid
boundaries with negligible impact on recall.

\subsection{Morpheme segmentation}
\label{sec:mainresults}
\begin{table*}[!h]
\centering
\caption{\textbf{Main results on the test set}: Wangchan reference, 2,000 documents, not used during
development. Boundary precision (P), recall (R) and F1, whitespace-adjacent boundaries excluded;
95\% confidence intervals (CI) from 1,000 resamples of the documents with replacement.}
\label{tab:main_results}
\setlength{\tabcolsep}{10pt}
\resizebox{\textwidth}{!}{%
\begin{tabular}{l l r r r r}
\toprule
\textbf{Type} & \textbf{System} & \textbf{P} $\uparrow$ & \textbf{R} $\uparrow$ & \textbf{F1} $\uparrow$ & \textbf{F1 95\% CI} \\
\midrule
\rowcolor{typhoonpurple!20}\multicolumn{6}{l}{\textit{\textbf{Word segmenters (BEST conventions)}}} \\
Dictionary & PyThaiNLP newmm & 0.962 & 0.687 & 0.802 & [0.800, 0.803] \\
Dictionary & nlpO3 (newmm in Rust) & 0.961 & 0.693 & 0.805 & [0.803, 0.807] \\
Neural & DeepCut & 0.972 & 0.875 & 0.921 & [0.919, 0.922] \\
Neural & AttaCut & 0.981 & 0.874 & 0.925 & [0.923, 0.926] \\
\midrule
\rowcolor{typhoonpurple!20}\multicolumn{6}{l}{\textit{\textbf{Simple morpheme baselines}}} \\
Rule & Syllable segmentation (PyThaiNLP) & 0.821 & \underline{0.991} & 0.898 & [0.896, 0.900] \\
Dictionary & Most-probable segmentation with a dictionary built from the LLM labels & 0.975 & 0.881 & 0.925 & [0.924, 0.927] \\
\midrule
\rowcolor{typhoonpurple!20}\multicolumn{6}{l}{\textit{\textbf{LLM}}} \\
LLM, 30B & Qwen3-30B-A3B two-pass segmenter (source of the training labels) & 0.975 & 0.948 & 0.962 & [0.960, 0.962] \\
\midrule
\rowcolor{typhoonpurple!20}\multicolumn{6}{l}{\textit{\textbf{Ours (1M parameters, CPU)}}} \\
BiLSTM & trained on uncorrected LLM labels & 0.977 & 0.948 & 0.962 & [0.961, 0.963] \\
BiLSTM & \textbf{\khammoon 0.1.0} & \textbf{0.992} & \textbf{0.993} & \textbf{0.992} & [0.992, 0.993] \\
\bottomrule
\end{tabular}}
\end{table*}

\Cref{tab:main_results} compares \khammoon with the
baselines on the held-out Wangchan test set.
\khammoon achieves the highest precision (0.992)
and recall (0.993), resulting in the best F1 (0.992).
Unlike the baselines, which tend to over-segment or
under-segment, \khammoon achieves high accuracy in
both boundary detection and boundary preservation.
We highlight three findings.

\begin{itemize}[nosep]
  \item \textbf{Word and syllable boundaries differ from
  morpheme boundaries.}
  Word segmenters achieve high precision (0.961--0.981)
  but lower recall (0.687--0.875), as they preserve
  compounds that the morpheme guideline splits.
  This reflects their different segmentation objective,
  rather than a failure at word segmentation.
  Conversely, syllable segmentation achieves high recall
  (0.991) but lower precision (0.821), since not every
  syllable is a morpheme.

  \item \textbf{Uncorrected LLM labels limit performance.}
  The LLM segmenter achieves 0.962 F1, with recall
  (0.948) lower than precision (0.975), consistent
  with its tendency to under-segment.
  Training the BiLSTM directly on these labels yields
  the same F1 (0.962), indicating that the architecture
  alone does not improve segmentation accuracy.

  \item \textbf{Corrected labels substantially improve
  segmentation.}
  \khammoon achieves 0.992 F1, an improvement of
  0.031 over the LLM segmenter (95\% CI:
  [0.030, 0.032]) and 0.068 over AttaCut
  (95\% CI: [0.066, 0.069]).
  This demonstrates the benefit of the correction
  and fine-tuning procedure.
\end{itemize}

\subsection{Speed and size}
\begin{table}[!b]
\centering
\caption{\textbf{Speed on one CPU thread}, in characters per second on dev-val.}
\label{tab:speed}
\begin{tabular}{l l r}
\toprule
\textbf{System} & \textbf{Output} & \textbf{Characters/s} $\uparrow$ \\
\midrule
nlpO3 & words & \textbf{2,475,000} \\
PyThaiNLP newmm & words & 547,000 \\
AttaCut & words & 44,500 \\
DeepCut & words & 1,700 \\
\midrule
\khammoon (ONNX Runtime) & morphemes & 32,100 \\
\khammoon (numpy) & morphemes & 12,900 \\
\bottomrule
\end{tabular}
\end{table}
\Cref{tab:speed} compares single-threaded CPU inference speed.
\khammoon processes 32,100 characters per second with ONNX
Runtime, approximately 1.4 times slower than AttaCut but
19 times faster than DeepCut. Its NumPy implementation
processes 12,900 characters per second. Dictionary-based
word segmenters are substantially faster, although they
produce word rather than morpheme boundaries.

\subsection{Downstream tasks}
\label{sec:downstream}
\begin{figure}[!tb]
\centering
\includegraphics[width=0.9\linewidth]{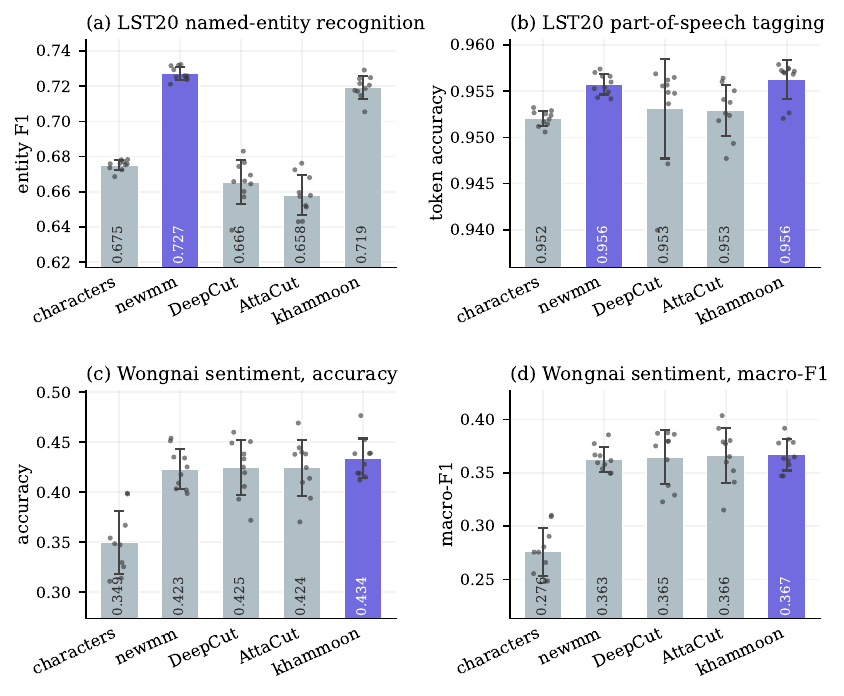}
\caption{\textbf{Downstream performance of taggers and classifiers trained on the output of each
segmenter.} (a)~Entity-level F1 of a BiLSTM tagger on LST20 named-entity recognition. (b)~Token
accuracy of the same tagger on LST20 part-of-speech tagging. (c)~Accuracy and (d)~macro-F1 of a
BiLSTM classifier on the Wongnai sentiment test set. Bars show the mean over the same ten random
seeds, error bars one standard deviation, and dots the individual seeds; the value at the foot of
each bar is the mean. In each panel the segmenters with the highest mean, to three decimals, are shown in purple. Across
(b)--(d) the seeds of khammoon and of the three word segmenters overlap; in (a) newmm leads
khammoon by 0.008 and both lead AttaCut and DeepCut by more than 0.05.
\Cref{tab:downstream} in Appendix~\ref{app:downstream} gives the numbers with paired
significance tests.}
\label{fig:downstream}
\end{figure}
\Cref{fig:downstream} compares the downstream performance
of \khammoon with other segmentation methods. Overall,
\khammoon performs comparably to word segmenters despite
producing finer-grained units.

\begin{itemize}[nosep]
  \item \textbf{Named-entity recognition.}
  \khammoon achieves an entity-level F1 of 0.719,
  outperforming AttaCut (0.658) and DeepCut (0.666)
  ($p<0.01$), but trailing newmm (0.727, $p=0.01$).
  This difference may reflect the closer alignment
  between newmm's word boundaries and LST20 annotations.

  \item \textbf{Part-of-speech tagging.}
  All segmenters perform similarly, with accuracy
  differing by at most 0.004.

  \item \textbf{Sentiment classification.}
  \khammoon achieves the highest mean accuracy (0.434)
  and macro-F1 (0.367), although its differences from
  word segmenters are not statistically significant
  ($p \geq 0.3$).
\end{itemize}

These results suggest that morpheme segmentation
provides finer-grained units without substantially
reducing downstream performance. Detailed results and
paired significance tests are reported in
\Cref{tab:downstream} (Appendix~\ref{app:downstream}).

\subsection{Error analysis}
On dev-val (768 sentences), \khammoon disagrees with the
reference in 227 regions: 142 with extra boundaries, 74 with
missing boundaries, and 11 with misplaced boundaries.
Some disagreements involve whitespace attachment, which
is excluded from our evaluation.

The remaining errors mainly involve lexical ambiguity.
\textbf{Proper names} are difficult because the model must
distinguish meaningful components from names that should
remain intact. For example, it fails to split \tht{สมปอง}
but incorrectly splits \tht{ธนะรัชต์}.
\textbf{Pali--Sanskrit compounds} are sometimes over-segmented
when their components also occur as independent Thai words,
as in \tht{ศาสน|จักร} and \tht{สังคม|วิทยา}.
Conversely, the model sometimes merges \textbf{adjacent
abbreviations} (e.g., \tht{บมจ.|กสท.}) and
\textbf{colloquial particles} (e.g., \tht{รึ|คะ}).
It also misses boundaries in some \textbf{rare compounds},
such as \tht{เผย|แพร่}, while occasionally splitting
single morphemes such as \tht{แซงแซว} (the bird name).

\section{Release}
\label{sec:release}
\paragraph{Model and library.}
We release \khammoon as a Python package
(\texttt{pip install khammoon}), with a segmentation API,
batch inference, and a command-line interface. The package
includes the 4.2\,MB model weights and supports inference through ONNX Runtime or NumPy.

\paragraph{Data and code.}
We release the training and evaluation data at
\texttt{typhoon-ai/khammoon-data}, including the original
and corrected LLM labels, human-labelled development and
test sets, and LLM predictions on both evaluation sets.
The release preserves source-text licenses and development
splits. Together with the training scripts, these resources
support reproducibility and provide a benchmark and training
data for future Thai morpheme segmentation research.

\paragraph{Licenses.} The code is released under Apache-2.0. The source texts carry NonCommercial
terms: BEST2009 is under CC BY-NC-SA 3.0 and WangchanThaiInstruct under CC BY-NC 4.0, with 5,014
training rows under CC BY-SA 4.0. The data therefore keep their per-row licenses, and the model
weights are released under CC BY-NC-SA 4.0.

\section{Limitations}
\textbf{Domain coverage.} The evaluation is conducted on instruction data spanning four domains. Because human-annotated morpheme benchmarks do not yet exist for informal web text or social media, the robustness of \khammoon in highly noisy or colloquial domains remains unverified.

\textbf{Lexical ambiguity and rare entities.} While \khammoon achieves high overall accuracy, error analysis reveals recurring challenges with rare proper names, complex transliterations, and Pali--Sanskrit compounds whose constituents overlap with independent native Thai words.

\textbf{Linguistic guideline scope.} Our segmentation reflects the formal criteria established in our guideline. Because Thai morphology productively incorporates novel foreign loanwords and colloquial coinages, future extensions to the taxonomy would require incremental re-annotation.

\textbf{Downstream scope.} Our downstream experiments assess segmentation utility on models trained from scratch without pretraining. Investigating the impact of morpheme units on large-scale pretrained language models, such as using morphemes to initialize subword tokenizer vocabularies, is an important direction left for future work.

\section{Conclusions}
We present \khammoon, a lightweight Thai morpheme
segmenter developed from a linguistically grounded
annotation guideline. Using this guideline, we construct
human-labelled evaluation sets and a large-scale
LLM-labelled training corpus, which we refine using
human annotations. The resulting model achieves a
boundary F1 of 0.992 on an out-of-domain test set
while being only 4\,MB in size and supporting
CPU inference. Despite producing finer-grained
segments than word segmenters, \khammoon achieves
comparable downstream performance on POS tagging,
NER, and sentiment classification. We publicly release
the annotation guideline, datasets, model, and library
to support future research on Thai segmentation.

\section*{Acknowledgments}
This research was supported by SCB DataX, Thailand, and the National Research Foundation, Singapore, under its National Large Language Models Funding Initiative. Any opinions, findings, conclusions, or recommendations expressed in this material are those of the authors and do not reflect the views of the funding organizations.

We thank NECTEC for the BEST corpus, VISTEC for WangchanThaiInstruct, the PyThaiNLP team, and the developers of AttaCut and DeepCut,
whose open resources this work builds on. Finally, we are deeply grateful to the members of the Typhoon team, past and present. Their ideas,
their work and their shared commitment to Thai language technology made this work possible, and it belongs to them as much as to its authors. 

\bibliographystyle{plainnat}
\bibliography{refs}

\clearpage
\beginappendix
\section{The Annotation Guideline}
\label{app:guideline}
This appendix reproduces the annotation guideline, the file
\texttt{linguistics/linguistic\_guideline.md}, as used by the annotators and, as a prompt, by the LLM segmenter. Each worked example is shown as
input $\rightarrow$ output, with \textbar{} marking a boundary and a bare \textbar{}~\textbar{} a
whitespace token. The two \tht{ทำลาย} examples under Category~4-D depend on the meaning in context,
as Step~3.3 prescribes. The guideline is subject to one hard constraint: the tokens must concatenate back to the
input exactly; no character may be added, removed or changed.
{\small Apply these steps sequentially for every input.

\subsection*{Step 1: Structural-level segmentation}
Split the text by syntactic/functional chunks. Preserve whitespace exactly.

\begin{itemize}[nosep,leftmargin=1.2em]
\item \tht{แม่ฝากฉันซื้อปลา} \tht{แต่ตลาดปิด} $\rightarrow$ \tht{แม่}\textbar{}\allowbreak{}\tht{ฝาก}\textbar{}\allowbreak{}\tht{ฉัน}\textbar{}\allowbreak{}\tht{ซื้อ}\textbar{}\allowbreak{}\tht{ปลา}\textbar{}\allowbreak{} \textbar{}\allowbreak{}\tht{แต่}\textbar{}\allowbreak{}\tht{ตลาด}\textbar{}\allowbreak{}\tht{ปิด}
\item \tht{ลลิษา} \tht{มโนบาลเป็นไอดอลสาวชาวไทย} $\rightarrow$ \tht{ลลิษา}\textbar{}\allowbreak{} \textbar{}\allowbreak{}\tht{มโนบาล}\textbar{}\allowbreak{}\tht{เป็น}\textbar{}\allowbreak{}\tht{ไอดอล}\textbar{}\allowbreak{}\tht{สาว}\textbar{}\allowbreak{}\tht{ชาว}\textbar{}\allowbreak{}\tht{ไทย}
\item \tht{ว้าย} \tht{นั่นใคร} $\rightarrow$ \tht{ว้าย}\textbar{}\allowbreak{} \textbar{}\allowbreak{}\tht{นั่น}\textbar{}\allowbreak{}\tht{ใคร}
\end{itemize}
\subsection*{Step 2: Non-Thai / punctuation / symbol handling}
\begin{itemize}[nosep,leftmargin=1.2em]
\item \textbf{KEEP (Do not split):}
\begin{itemize}[nosep,leftmargin=1.2em]
\item Technical data (URLs, file paths, emails, tags, filenames)
\begin{itemize}[nosep,leftmargin=1.2em]
\item \tht{ไปที่} www.google.com \tht{เลย} $\rightarrow$ \tht{ไป}\textbar{}\allowbreak{}\tht{ที่}\textbar{}\allowbreak{} \textbar{}\allowbreak{}www.google.com\textbar{}\allowbreak{} \textbar{}\allowbreak{}\tht{เลย}
\item \tht{ชื่อไฟล์} screenshot\_2.png \tht{ค่ะ} $\rightarrow$ \tht{ชื่อ}\textbar{}\allowbreak{}\tht{ไฟล์}\textbar{}\allowbreak{} \textbar{}\allowbreak{}screenshot\_2.png\textbar{}\allowbreak{} \textbar{}\allowbreak{}\tht{ค่ะ}
\item \tht{ส่งมาที่} mike@abc.co.uk \tht{ครับ} $\rightarrow$ \tht{ส่ง}\textbar{}\allowbreak{}\tht{มา}\textbar{}\allowbreak{}\tht{ที่}\textbar{}\allowbreak{} \textbar{}\allowbreak{}mike@abc.co.uk\textbar{}\allowbreak{} \textbar{}\allowbreak{}\tht{ครับ}
\end{itemize}
\item Numeric values with commas or decimal points
\begin{itemize}[nosep,leftmargin=1.2em]
\item \tht{เงิน} 2,552.24 \tht{พันล้านบาท} $\rightarrow$ \tht{เงิน}\textbar{}\allowbreak{} \textbar{}\allowbreak{}2,552.24\textbar{}\allowbreak{} \textbar{}\allowbreak{}\tht{พัน}\textbar{}\allowbreak{}\tht{ล้าน}\textbar{}\allowbreak{}\tht{บาท}
\item \tht{จำนวนยอด} 18,603 \tht{บาท} $\rightarrow$ \tht{จำนวน}\textbar{}\allowbreak{}\tht{ยอด}\textbar{}\allowbreak{} \textbar{}\allowbreak{}18,603\textbar{}\allowbreak{} \textbar{}\allowbreak{}\tht{บาท}
\item \tht{โรงเรียนอยู่ห่างออกไป} 720.5 \tht{เมตร} $\rightarrow$ \tht{โรง}\textbar{}\allowbreak{}\tht{เรียน}\textbar{}\allowbreak{}\tht{อยู่}\textbar{}\allowbreak{}\tht{ห่าง}\textbar{}\allowbreak{}\tht{ออก}\textbar{}\allowbreak{}\tht{ไป}\textbar{}\allowbreak{} \textbar{}\allowbreak{}720.5\textbar{}\allowbreak{} \textbar{}\allowbreak{}\tht{เมตร}
\end{itemize}
\item Section/heading numbers
\begin{itemize}[nosep,leftmargin=1.2em]
\item 1. \tht{นกนางนวล} $\rightarrow$ 1.\textbar{}\allowbreak{} \textbar{}\allowbreak{}\tht{นก}\textbar{}\allowbreak{}\tht{นาง}\textbar{}\allowbreak{}\tht{นวล}
\item 3) \tht{นางสมร} $\rightarrow$ 3)\textbar{}\allowbreak{} \textbar{}\allowbreak{}\tht{นาง}\textbar{}\allowbreak{}\tht{สมร}
\item \tht{บทความที่} (\tht{๒}) \tht{ระบุว่า} $\rightarrow$ \tht{บท}\textbar{}\allowbreak{}\tht{ความ}\textbar{}\allowbreak{}\tht{ที่}\textbar{}\allowbreak{} \textbar{}\allowbreak{}(\tht{๒})\textbar{}\allowbreak{} \textbar{}\allowbreak{}\tht{ระบุ}\textbar{}\allowbreak{}\tht{ว่า}
\end{itemize}
\item Chemical compounds
\begin{itemize}[nosep,leftmargin=1.2em]
\item Ca(OH)2 + CO2 $\rightarrow$ Ca(OH)2\textbar{}\allowbreak{} \textbar{}\allowbreak{}+\textbar{}\allowbreak{} \textbar{}\allowbreak{}CO2
\end{itemize}
\end{itemize}
\item \textbf{SPLIT (Separate into individual tokens):}
\begin{itemize}[nosep,leftmargin=1.2em]
\item Foreign-language text (split on whitespace)
\begin{itemize}[nosep,leftmargin=1.2em]
\item into the borderland $\rightarrow$ into\textbar{}\allowbreak{} \textbar{}\allowbreak{}the\textbar{}\allowbreak{} \textbar{}\allowbreak{}borderland
\item the 4th generation $\rightarrow$ the\textbar{}\allowbreak{} \textbar{}\allowbreak{}4th\textbar{}\allowbreak{} \textbar{}\allowbreak{}generation
\item People's Alliance for Democracy $\rightarrow$ People's\textbar{}\allowbreak{} \textbar{}\allowbreak{}Alliance\textbar{}\allowbreak{} \textbar{}\allowbreak{}for\textbar{}\allowbreak{} \textbar{}\allowbreak{}Democracy
\item That’s bad $\rightarrow$ That’s\textbar{}\allowbreak{} \textbar{}\allowbreak{}bad
\end{itemize}
\item Punctuation (parentheses, quotes)
\begin{itemize}[nosep,leftmargin=1.2em]
\item \tht{บดินทรเดชา} (\tht{สิงห์} \tht{สิงหเสนี}) $\rightarrow$ \tht{บดินทรเดชา}\textbar{}\allowbreak{} \textbar{}\allowbreak{}(\textbar{}\allowbreak{}\tht{สิงห์}\textbar{}\allowbreak{} \textbar{}\allowbreak{}\tht{สิงหเสนี}\textbar{}\allowbreak{})
\item \tht{เขาร้องว่า} “\tht{ช่วยด้วย}” $\rightarrow$ \tht{เขา}\textbar{}\allowbreak{}\tht{ร้อง}\textbar{}\allowbreak{}\tht{ว่า}\textbar{}\allowbreak{} \textbar{}\allowbreak{}“\textbar{}\allowbreak{}\tht{ช่วย}\textbar{}\allowbreak{}\tht{ด้วย}\textbar{}\allowbreak{}”
\end{itemize}
\item Symbols (\%, \#, etc.) separated from adjacent text
\begin{itemize}[nosep,leftmargin=1.2em]
\item \tht{มีขยะเพิ่มเป็น} 25.50\% $\rightarrow$ \tht{มี}\textbar{}\allowbreak{}\tht{ขยะ}\textbar{}\allowbreak{}\tht{เพิ่ม}\textbar{}\allowbreak{}\tht{เป็น}\textbar{}\allowbreak{} \textbar{}\allowbreak{}25.50\textbar{}\allowbreak{}\%
\item \#\tht{วันหยุด} \#\tht{ทะเล} $\rightarrow$ \#\textbar{}\allowbreak{}\tht{วัน}\textbar{}\allowbreak{}\tht{หยุด}\textbar{}\allowbreak{} \textbar{}\allowbreak{}\#\textbar{}\allowbreak{}\tht{ทะเล}
\item \tht{ไม่มีอีกแล้ว}.... \tht{เงินเก็บ} $\rightarrow$ \tht{ไม่}\textbar{}\allowbreak{}\tht{มี}\textbar{}\allowbreak{}\tht{อีก}\textbar{}\allowbreak{}\tht{แล้ว}\textbar{}\allowbreak{}.\textbar{}\allowbreak{}.\textbar{}\allowbreak{}.\textbar{}\allowbreak{}.\textbar{}\allowbreak{} \textbar{}\allowbreak{}\tht{เงิน}\textbar{}\allowbreak{}\tht{เก็บ}
\item \tht{บริษัท} Yahoo! \tht{มีกำไร} \$300,000-400,000 $\rightarrow$ \tht{บริษัท}\textbar{}\allowbreak{} \textbar{}\allowbreak{}Yahoo\textbar{}\allowbreak{}!\textbar{}\allowbreak{} \textbar{}\allowbreak{}\tht{มี}\textbar{}\allowbreak{}\tht{กำไร}\textbar{}\allowbreak{} \textbar{}\allowbreak{}\$\textbar{}\allowbreak{}300,000\textbar{}\allowbreak{}-\textbar{}\allowbreak{}400,000
\item \tht{ลดราคา} 25\% $\rightarrow$ \tht{ลด}\textbar{}\allowbreak{}\tht{ราคา}\textbar{}\allowbreak{} \textbar{}\allowbreak{}25\textbar{}\allowbreak{}\%
\end{itemize}
\item Mathematical/chemical symbols (+, -, =, \~{}, \^{}, --->, etc.) separated from adjacent text
\begin{itemize}[nosep,leftmargin=1.2em]
\item Ca(OH)2 + CO2 ---> CaCO3 + H2O + heat $\rightarrow$ Ca(OH)2\textbar{}\allowbreak{} \textbar{}\allowbreak{}+\textbar{}\allowbreak{} \textbar{}\allowbreak{}CO2\textbar{}\allowbreak{} \textbar{}\allowbreak{}--->\textbar{}\allowbreak{} \textbar{}\allowbreak{}CaCO3\textbar{}\allowbreak{} \textbar{}\allowbreak{}+\textbar{}\allowbreak{} \textbar{}\allowbreak{}H2O\textbar{}\allowbreak{} \textbar{}\allowbreak{}+\textbar{}\allowbreak{} \textbar{}\allowbreak{}heat
\item {[(18*67)+15]}-17\^{}2 = \~{}42.22 $\rightarrow$ [\textbar{}\allowbreak{}(\textbar{}\allowbreak{}18\textbar{}\allowbreak{}*\textbar{}\allowbreak{}67\textbar{}\allowbreak{})\textbar{}\allowbreak{}+\textbar{}\allowbreak{}15\textbar{}\allowbreak{}]\textbar{}\allowbreak{}-\textbar{}\allowbreak{}17\textbar{}\allowbreak{}\^{}\textbar{}\allowbreak{}2\textbar{}\allowbreak{} \textbar{}\allowbreak{}=\textbar{}\allowbreak{} \textbar{}\allowbreak{}\~{}\textbar{}\allowbreak{}42.22
\end{itemize}
\end{itemize}
\end{itemize}
\subsection*{Step 3: Thai-character handling}
\begin{itemize}[nosep,leftmargin=1.2em]
\item \textbf{3.1 Assume correct orthography:} Segment based on intended valid spelling.
\begin{itemize}[nosep,leftmargin=1.2em]
\item \tht{ศรีษะมาร} $\rightarrow$ \tht{ศรีษะ}\textbar{}\allowbreak{}\tht{มาร}
\item \tht{พะยักพะเยิด} $\rightarrow$ \tht{พะยัก}\textbar{}\allowbreak{}\tht{พะเยิด}
\item \tht{กะหรอมกะแหรม} $\rightarrow$ \tht{กะ}\textbar{}\allowbreak{}\tht{หรอม}\textbar{}\allowbreak{}\tht{กะ}\textbar{}\allowbreak{}\tht{แหรม}
\item \tht{กระเพราผัดไก่} $\rightarrow$ \tht{กระเพรา}\textbar{}\allowbreak{}\tht{ผัด}\textbar{}\allowbreak{}\tht{ไก่}
\end{itemize}
\item \textbf{3.2 Abbreviations \& Punctuation:}
\begin{itemize}[nosep,leftmargin=1.2em]
\item Single entity rank/unit/place -> KEEP. Stacked distinct credentials -> SPLIT.
\begin{itemize}[nosep,leftmargin=1.2em]
\item \tht{ผบ}.\tht{ทบ}.\tht{สั่งการว่า} $\rightarrow$ \tht{ผบ}.\tht{ทบ}.\textbar{}\allowbreak{}\tht{สั่ง}\textbar{}\allowbreak{}\tht{การ}\textbar{}\allowbreak{}\tht{ว่า}
\item \tht{คณะอนุกก}. $\rightarrow$ \tht{คณะ}\textbar{}\allowbreak{}\tht{อนุกก}.
\item \tht{ผบ}.\tht{ภ}.\tht{อ}.\tht{คง} \tht{กล่าวว่า} $\rightarrow$ \tht{ผบ}.\tht{ภ}.\textbar{}\allowbreak{}\tht{อ}.\textbar{}\allowbreak{}\tht{คง}\textbar{}\allowbreak{} \textbar{}\allowbreak{}\tht{กล่าว}\textbar{}\allowbreak{}\tht{ว่า}
\item \tht{รศ}.\tht{ดร}.\tht{พญ}.\tht{วีณา} $\rightarrow$ \tht{รศ}.\textbar{}\allowbreak{}\tht{ดร}.\textbar{}\allowbreak{}\tht{พญ}.\textbar{}\allowbreak{}\tht{วีณา}
\item \tht{พ}.\tht{ศ}. 2569 $\rightarrow$ \tht{พ}.\tht{ศ}.\textbar{}\allowbreak{} \textbar{}\allowbreak{}2569
\end{itemize}
\item Ellipsis (\tht{ฯ}, \tht{ฯลฯ}) -> SPLIT (Exception: \tht{ฯพณฯ} is KEEP).
\begin{itemize}[nosep,leftmargin=1.2em]
\item \tht{กรุงเทพฯ} $\rightarrow$ \tht{กรุง}\textbar{}\allowbreak{}\tht{เทพ}\textbar{}\allowbreak{}\tht{ฯ}
\item \tht{กล้วย} \tht{ส้ม} \tht{ฯลฯ} $\rightarrow$ \tht{กล้วย}\textbar{}\allowbreak{} \textbar{}\allowbreak{}\tht{ส้ม}\textbar{}\allowbreak{} \textbar{}\allowbreak{}\tht{ฯลฯ}
\item \tht{รอต้อนรับ} \tht{ฯพณฯ} \tht{นายกรัฐมนตรี} $\rightarrow$ \tht{รอ}\textbar{}\allowbreak{}\tht{ต้อน}\textbar{}\allowbreak{}\tht{รับ}\textbar{}\allowbreak{} \textbar{}\allowbreak{}\tht{ฯพณฯ}\textbar{}\allowbreak{} \textbar{}\allowbreak{}\tht{นายก}\textbar{}\allowbreak{}\tht{รัฐมนตรี}
\end{itemize}
\item \tht{ไม้ยมก} (\tht{ๆ}) -> SPLIT.
\begin{itemize}[nosep,leftmargin=1.2em]
\item \tht{รักมากๆๆ} $\rightarrow$ \tht{รัก}\textbar{}\allowbreak{}\tht{มาก}\textbar{}\allowbreak{}\tht{ๆ}\textbar{}\allowbreak{}\tht{ๆ}
\end{itemize}
\end{itemize}
\item \textbf{3.3 Context Resolution:} Read the full sentence to determine meaning BEFORE Step 4. Tests in Step 4 MUST use this contextually determined meaning.
\begin{itemize}[nosep,leftmargin=1.2em]
\item \tht{ถ้าจะให้ตัดเสื้อก็วัดรอบอกมา} $\rightarrow$ \tht{ถ้า}\textbar{}\allowbreak{}\tht{จะ}\textbar{}\allowbreak{}\tht{ให้}\textbar{}\allowbreak{}\tht{ตัด}\textbar{}\allowbreak{}\tht{เสื้อ}\textbar{}\allowbreak{}\tht{ก็}\textbar{}\allowbreak{}\tht{วัด}\textbar{}\allowbreak{}\tht{รอบ}\textbar{}\allowbreak{}\tht{อก}\textbar{}\allowbreak{}\tht{มา}
\item \tht{คงไม่มีใครจะมารอบอกเธอหรอก} $\rightarrow$ \tht{คง}\textbar{}\allowbreak{}\tht{ไม่}\textbar{}\allowbreak{}\tht{มี}\textbar{}\allowbreak{}\tht{ใคร}\textbar{}\allowbreak{}\tht{จะ}\textbar{}\allowbreak{}\tht{มา}\textbar{}\allowbreak{}\tht{รอ}\textbar{}\allowbreak{}\tht{บอก}\textbar{}\allowbreak{}\tht{เธอ}\textbar{}\allowbreak{}\tht{หรอก}
\item \tht{พายุพัดเข้าฝั่งทำลายบ้านเรือน} $\rightarrow$ \tht{พายุ}\textbar{}\allowbreak{}\tht{พัด}\textbar{}\allowbreak{}\tht{เข้า}\textbar{}\allowbreak{}\tht{ฝั่ง}\textbar{}\allowbreak{}\tht{ทำลาย}\textbar{}\allowbreak{}\tht{บ้าน}\textbar{}\allowbreak{}\tht{เรือน}
\item \tht{ตัดกระดาษแล้วทำลายเป็นรูปดาว} $\rightarrow$ \tht{ตัด}\textbar{}\allowbreak{}\tht{กระดาษ}\textbar{}\allowbreak{}\tht{แล้ว}\textbar{}\allowbreak{}\tht{ทำ}\textbar{}\allowbreak{}\tht{ลาย}\textbar{}\allowbreak{}\tht{เป็น}\textbar{}\allowbreak{}\tht{รูป}\textbar{}\allowbreak{}\tht{ดาว}
\end{itemize}
\end{itemize}
\subsection*{Step 4: Decide monomorphemic units (\tht{คำมูล}) vs. compound words}
Evaluate each remaining chunk. Stop at the first category that applies.

\begin{itemize}[nosep,leftmargin=1.2em]
\item \textbf{Category 1: Mixed origins -> SPLIT}
\begin{itemize}[nosep,leftmargin=1.2em]
\item \tht{เคมีภัณฑ์} $\rightarrow$ \tht{เคมี}\textbar{}\allowbreak{}\tht{ภัณฑ์}
\item \tht{พระเก้าอี้} $\rightarrow$ \tht{พระ}\textbar{}\allowbreak{}\tht{เก้าอี้}
\item \tht{ราชวัง} $\rightarrow$ \tht{ราช}\textbar{}\allowbreak{}\tht{วัง}
\item \tht{ผลไม้} $\rightarrow$ \tht{ผล}\textbar{}\allowbreak{}\tht{ไม้}
\item \tht{ซุปกิมจิ} $\rightarrow$ \tht{ซุป}\textbar{}\allowbreak{}\tht{กิมจิ}
\item \tht{คุณค่าของราชวังเก่า} $\rightarrow$ \tht{คุณ}\textbar{}\allowbreak{}\tht{ค่า}\textbar{}\allowbreak{}\tht{ของ}\textbar{}\allowbreak{}\tht{ราช}\textbar{}\allowbreak{}\tht{วัง}\textbar{}\allowbreak{}\tht{เก่า}
\item \tht{อุตสาหกรรมเคมีภัณฑ์} $\rightarrow$ \tht{อุตสาหกรรม}\textbar{}\allowbreak{}\tht{เคมี}\textbar{}\allowbreak{}\tht{ภัณฑ์}
\item \tht{เวชระเบียน} $\rightarrow$ \tht{เวช}\textbar{}\allowbreak{}\tht{ระเบียน}
\item \tht{พลเมือง} $\rightarrow$ \tht{พล}\textbar{}\allowbreak{}\tht{เมือง}
\end{itemize}
\item \textbf{Category 2: All Pali-Sanskrit}
\begin{itemize}[nosep,leftmargin=1.2em]
\item True \tht{สมาส}-\tht{สนธิ} (Meaning flows back->front): KEEP.
\begin{itemize}[nosep,leftmargin=1.2em]
\item \tht{ประวัติศาสตร์} $\rightarrow$ \tht{ประวัติศาสตร์}
\item \tht{แพทยสภา} $\rightarrow$ \tht{แพทยสภา}
\item \tht{พุทธศาสนา} $\rightarrow$ \tht{พุทธศาสนา}
\item \tht{มหาบัณฑิต} $\rightarrow$ \tht{มหาบัณฑิต}
\end{itemize}
\item Thai-syntax order (front->back) or \tht{คำซ้อน}: SPLIT.
\begin{itemize}[nosep,leftmargin=1.2em]
\item \tht{ศาสนาพุทธ} $\rightarrow$ \tht{ศาสนา}\textbar{}\allowbreak{}\tht{พุทธ}
\item \tht{ฤทธิ์ณรงค์} $\rightarrow$ \tht{ฤทธิ์}\textbar{}\allowbreak{}\tht{ณรงค์}
\item \tht{ประชาชน} $\rightarrow$ \tht{ประชา}\textbar{}\allowbreak{}\tht{ชน}
\end{itemize}
\end{itemize}
\item \textbf{Category 3: Loanwords / transliterations}
\begin{itemize}[nosep,leftmargin=1.2em]
\item Interpreted back->front: KEEP.
\begin{itemize}[nosep,leftmargin=1.2em]
\item \tht{วิปครีม} $\rightarrow$ \tht{วิปครีม}
\item \tht{คอนกรีตบล็อก} $\rightarrow$ \tht{คอนกรีตบล็อก}
\end{itemize}
\item Interpreted front->back: SPLIT.
\begin{itemize}[nosep,leftmargin=1.2em]
\item \tht{คุกกี้ช็อกโกแลต} $\rightarrow$ \tht{คุกกี้}\textbar{}\allowbreak{}\tht{ช็อกโกแลต}
\item \tht{เค้กวานิลลา} $\rightarrow$ \tht{เค้ก}\textbar{}\allowbreak{}\tht{วานิลลา}
\end{itemize}
\end{itemize}
\item \textbf{Category 4: Native Thai words}
\begin{itemize}[nosep,leftmargin=1.2em]
\item \textbf{4-A Sound Reduction (\tht{กร่อนเสียง}): KEEP.}
\begin{itemize}[nosep,leftmargin=1.2em]
\item \tht{ตะวัน} $\rightarrow$ \tht{ตะวัน}
\item \tht{มะม่วง} $\rightarrow$ \tht{มะม่วง}
\item \tht{มะพร้าว} $\rightarrow$ \tht{มะพร้าว}
\item \tht{กินมะม่วงกะมะพร้าว} $\rightarrow$ \tht{กิน}\textbar{}\allowbreak{}\tht{มะม่วง}\textbar{}\allowbreak{}\tht{กะ}\textbar{}\allowbreak{}\tht{มะพร้าว}
\end{itemize}
\item \textbf{4-B Meaningless affix syllables: SPLIT (unless it is a fused prefix like \tht{ระ}-, \tht{กระ}-, \tht{ประ}-).}
\begin{itemize}[nosep,leftmargin=1.2em]
\item \tht{ตาเตอ} $\rightarrow$ \tht{ตา}\textbar{}\allowbreak{}\tht{เตอ}
\item \tht{ร้อนเริ้น} $\rightarrow$ \tht{ร้อน}\textbar{}\allowbreak{}\tht{เริ้น}
\item \tht{กระป๋งกระเป๋า} $\rightarrow$ \tht{กระป๋ง}\textbar{}\allowbreak{}\tht{กระเป๋า}
\item \tht{พะรุงพะรัง} $\rightarrow$ \tht{พะ}\textbar{}\allowbreak{}\tht{รุง}\textbar{}\allowbreak{}\tht{พะ}\textbar{}\allowbreak{}\tht{รัง}
\item \tht{กะหนุงกะหนิง} $\rightarrow$ \tht{กะ}\textbar{}\allowbreak{}\tht{หนุง}\textbar{}\allowbreak{}\tht{กะ}\textbar{}\allowbreak{}\tht{หนิง}
\item \tht{ดี๊ดี} $\rightarrow$ \tht{ดี๊}\textbar{}\allowbreak{}\tht{ดี}
\item \tht{อากาศร้อนเริ้น} $\rightarrow$ \tht{อากาศ}\textbar{}\allowbreak{}\tht{ร้อน}\textbar{}\allowbreak{}\tht{เริ้น}
\item \tht{ของวางพะรุงพะรัง} $\rightarrow$ \tht{ของ}\textbar{}\allowbreak{}\tht{วาง}\textbar{}\allowbreak{}\tht{พะ}\textbar{}\allowbreak{}\tht{รุง}\textbar{}\allowbreak{}\tht{พะ}\textbar{}\allowbreak{}\tht{รัง}
\item \tht{กะหนุงกะหนิงกันทั้งวัน} $\rightarrow$ \tht{กะ}\textbar{}\allowbreak{}\tht{หนุง}\textbar{}\allowbreak{}\tht{กะ}\textbar{}\allowbreak{}\tht{หนิง}\textbar{}\allowbreak{}\tht{กัน}\textbar{}\allowbreak{}\tht{ทั้ง}\textbar{}\allowbreak{}\tht{วัน}
\end{itemize}
\item \textit{(Fused Prefixes to KEEP)}
\begin{itemize}[nosep,leftmargin=1.2em]
\item \tht{ระรื่น} $\rightarrow$ \tht{ระรื่น}
\item \tht{กระโดด} $\rightarrow$ \tht{กระโดด}
\item \tht{ประเดี๋ยว} $\rightarrow$ \tht{ประเดี๋ยว}
\item \tht{กระผม} $\rightarrow$ \tht{กระผม}
\item \tht{ประเดี๋ยวกระผมนำมาให้} $\rightarrow$ \tht{ประเดี๋ยว}\textbar{}\allowbreak{}\tht{กระผม}\textbar{}\allowbreak{}\tht{นำ}\textbar{}\allowbreak{}\tht{มา}\textbar{}\allowbreak{}\tht{ให้}
\end{itemize}
\item \textbf{4-C Onomatopoeia (\tht{คำเลียนเสียงธรรมชาติ})}
\\ Base ABAB -> KEEP. Explicit doubling/interleaved -> SPLIT.
\begin{itemize}[nosep,leftmargin=1.2em]
\item \tht{คิกคัก} $\rightarrow$ \tht{คิกคัก}
\item \tht{ฉึกฉัก} $\rightarrow$ \tht{ฉึกฉัก}
\item \tht{ยุกยิก} $\rightarrow$ \tht{ยุกยิก}
\item \tht{ตึกตัก} $\rightarrow$ \tht{ตึกตัก}
\item \tht{พึบพับ} $\rightarrow$ \tht{พึบพับ}
\item \tht{คิกคิกคักคัก} $\rightarrow$ \tht{คิก}\textbar{}\allowbreak{}\tht{คิก}\textbar{}\allowbreak{}\tht{คัก}\textbar{}\allowbreak{}\tht{คัก}
\item \tht{พึบพึบพับพับ} $\rightarrow$ \tht{พึบ}\textbar{}\allowbreak{}\tht{พึบ}\textbar{}\allowbreak{}\tht{พับ}\textbar{}\allowbreak{}\tht{พับ}
\item \tht{ฉึกฉึกฉักฉัก} $\rightarrow$ \tht{ฉึก}\textbar{}\allowbreak{}\tht{ฉึก}\textbar{}\allowbreak{}\tht{ฉัก}\textbar{}\allowbreak{}\tht{ฉัก}
\item \tht{ยุกยุกยิกยิก} $\rightarrow$ \tht{ยุก}\textbar{}\allowbreak{}\tht{ยุก}\textbar{}\allowbreak{}\tht{ยิก}\textbar{}\allowbreak{}\tht{ยิก}
\item \tht{ตึกตึกตักตัก} $\rightarrow$ \tht{ตึก}\textbar{}\allowbreak{}\tht{ตึก}\textbar{}\allowbreak{}\tht{ตัก}\textbar{}\allowbreak{}\tht{ตัก}
\item \tht{กระฉึกกระฉาก} $\rightarrow$ \tht{กระ}\textbar{}\allowbreak{}\tht{ฉึก}\textbar{}\allowbreak{}\tht{กระ}\textbar{}\allowbreak{}\tht{ฉาก}
\end{itemize}
\item \textbf{4-D Meaning substitution / insertion test (using Step 3.3 context)}
\\ If meaningful word can be inserted or substituted -> SPLIT. Else -> KEEP.
\begin{itemize}[nosep,leftmargin=1.2em]
\item \tht{รถไฟ} $\rightarrow$ \tht{รถ}\textbar{}\allowbreak{}\tht{ไฟ}
\item \tht{ทิศตะวันออก} $\rightarrow$ \tht{ทิศ}\textbar{}\allowbreak{}\tht{ตะวัน}\textbar{}\allowbreak{}\tht{ออก}
\item \tht{มือถือ} $\rightarrow$ \tht{มือ}\textbar{}\allowbreak{}\tht{ถือ}
\item \tht{ทำลาย} (meaning: to make patterns) $\rightarrow$ \tht{ทำ}\textbar{}\allowbreak{}\tht{ลาย}
\item \tht{ทำลาย} (meaning: to destroy) $\rightarrow$ \tht{ทำลาย}
\item \tht{นาฬิกา} $\rightarrow$ \tht{นาฬิกา}
\item \tht{กระตือรือร้น} $\rightarrow$ \tht{กระตือรือร้น}
\end{itemize}
\end{itemize}
\item \textbf{Category 5: Unknown origin or persistent uncertainty -> Pronunciation test}
\\ If pronunciation unchanged -> SPLIT. If pronunciation changes -> KEEP.
\begin{itemize}[nosep,leftmargin=1.2em]
\item \tht{ขะมักเขม้น} $\rightarrow$ \tht{ขะมัก}\textbar{}\allowbreak{}\tht{เขม้น}
\item \tht{จิ้มลิ้ม} $\rightarrow$ \tht{จิ้ม}\textbar{}\allowbreak{}\tht{ลิ้ม}
\item \tht{จุกจิก} $\rightarrow$ \tht{จุก}\textbar{}\allowbreak{}\tht{จิก}
\item \tht{พระสงฆ์} $\rightarrow$ \tht{พระ}\textbar{}\allowbreak{}\tht{สงฆ์}
\item \tht{ภาคภูมิ} $\rightarrow$ \tht{ภาค}\textbar{}\allowbreak{}\tht{ภูมิ}
\item \tht{อนึ่ง} $\rightarrow$ \tht{อนึ่ง}
\end{itemize}
\end{itemize}
\subsection*{Additional worked examples}
\begin{itemize}[nosep,leftmargin=1.2em]
\item \tht{โรงเรียนสอนคนตาบอด} $\rightarrow$ \tht{โรง}\textbar{}\allowbreak{}\tht{เรียน}\textbar{}\allowbreak{}\tht{สอน}\textbar{}\allowbreak{}\tht{คน}\textbar{}\allowbreak{}\tht{ตา}\textbar{}\allowbreak{}\tht{บอด}
\item \tht{พลเมืองดีศรีสังคม} $\rightarrow$ \tht{พล}\textbar{}\allowbreak{}\tht{เมือง}\textbar{}\allowbreak{}\tht{ดี}\textbar{}\allowbreak{}\tht{ศรี}\textbar{}\allowbreak{}\tht{สังคม}
\item \tht{รัฐบาลประกาศนโยบาย} $\rightarrow$ \tht{รัฐบาล}\textbar{}\allowbreak{}\tht{ประกาศ}\textbar{}\allowbreak{}\tht{นโยบาย}
\item \tht{นายสมชายทำงานที่ปารีส} $\rightarrow$ \tht{นาย}\textbar{}\allowbreak{}\tht{สม}\textbar{}\allowbreak{}\tht{ชาย}\textbar{}\allowbreak{}\tht{ทำ}\textbar{}\allowbreak{}\tht{งาน}\textbar{}\allowbreak{}\tht{ที่}\textbar{}\allowbreak{}\tht{ปารีส}
\item \tht{กลุ่มองค์กรต่างๆ} \tht{ในสังคมนานาชาติ} $\rightarrow$ \tht{กลุ่ม}\textbar{}\allowbreak{}\tht{องค์กร}\textbar{}\allowbreak{}\tht{ต่าง}\textbar{}\allowbreak{}\tht{ๆ}\textbar{}\allowbreak{} \textbar{}\allowbreak{}\tht{ใน}\textbar{}\allowbreak{}\tht{สังคม}\textbar{}\allowbreak{}\tht{นานาชาติ}
\item \tht{มหาวิทยาลัยเชียงใหม่} $\rightarrow$ \tht{มหาวิทยาลัย}\textbar{}\allowbreak{}\tht{เชียง}\textbar{}\allowbreak{}\tht{ใหม่}
\item \tht{คุณค่าของชุมชนราชวังเก่า} $\rightarrow$ \tht{คุณ}\textbar{}\allowbreak{}\tht{ค่า}\textbar{}\allowbreak{}\tht{ของ}\textbar{}\allowbreak{}\tht{ชุม}\textbar{}\allowbreak{}\tht{ชน}\textbar{}\allowbreak{}\tht{ราช}\textbar{}\allowbreak{}\tht{วัง}\textbar{}\allowbreak{}\tht{เก่า}
\item \tht{อุตสาหกรรมไฟฟ้าในประเทศไทย} $\rightarrow$ \tht{อุตสาหกรรม}\textbar{}\allowbreak{}\tht{ไฟ}\textbar{}\allowbreak{}\tht{ฟ้า}\textbar{}\allowbreak{}\tht{ใน}\textbar{}\allowbreak{}\tht{ประเทศ}\textbar{}\allowbreak{}\tht{ไทย}
\item \tht{ผกก}.\tht{ภ}.\tht{อ}.\tht{คง} \tht{กล่าว} $\rightarrow$ \tht{ผกก}.\tht{ภ}.\textbar{}\allowbreak{}\tht{อ}.\textbar{}\allowbreak{}\tht{คง}\textbar{}\allowbreak{} \textbar{}\allowbreak{}\tht{กล่าว}
\item \tht{พล}.\tht{ต}.\tht{อ}.\tht{จงรัก} \tht{รอง} \tht{ผบ}.\tht{ตร}.(\tht{ปป}.3) $\rightarrow$ \tht{พล}.\tht{ต}.\tht{อ}.\textbar{}\allowbreak{}\tht{จง}\textbar{}\allowbreak{}\tht{รัก}\textbar{}\allowbreak{} \textbar{}\allowbreak{}\tht{รอง}\textbar{}\allowbreak{} \textbar{}\allowbreak{}\tht{ผบ}.\tht{ตร}.\textbar{}\allowbreak{}(\textbar{}\allowbreak{}\tht{ปป}.\textbar{}\allowbreak{}3\textbar{}\allowbreak{})
\item \tht{ผบ}.\tht{สน}.\tht{ทองหล่อบุกจับกุมผศ}.\tht{ดร}.\tht{วัฒนแสงประเสริฐ} $\rightarrow$ \tht{ผบ}.\textbar{}\allowbreak{}\tht{สน}.\textbar{}\allowbreak{}\tht{ทอง}\textbar{}\allowbreak{}\tht{หล่อ}\textbar{}\allowbreak{}\tht{บุก}\textbar{}\allowbreak{}\tht{จับ}\textbar{}\allowbreak{}\tht{กุม}\textbar{}\allowbreak{}\tht{ผศ}.\textbar{}\allowbreak{}\tht{ดร}.\textbar{}\allowbreak{}\tht{วัฒน}\textbar{}\allowbreak{}\tht{แสง}\textbar{}\allowbreak{}\tht{ประเสริฐ}
\item \tht{ผบ}.\tht{สส}.\tht{พงส์เพ็ชรปฏิเสสเหตุการณ์ดังกล่าว} $\rightarrow$ \tht{ผบ}.\tht{สส}.\textbar{}\allowbreak{}\tht{พงส์}\textbar{}\allowbreak{}\tht{เพ็ชร}\textbar{}\allowbreak{}\tht{ปฏิเสส}\textbar{}\allowbreak{}\tht{เหตุ}\textbar{}\allowbreak{}\tht{การณ์}\textbar{}\allowbreak{}\tht{ดัง}\textbar{}\allowbreak{}\tht{กล่าว}
\item \tht{กป}.\tht{อพช}.\tht{ภาคอีสาน} 12-14/12/2559 $\rightarrow$ \tht{กป}.\tht{อพช}.\textbar{}\allowbreak{}\tht{ภาค}\textbar{}\allowbreak{}\tht{อีสาน}\textbar{}\allowbreak{} \textbar{}\allowbreak{}12\textbar{}\allowbreak{}-\textbar{}\allowbreak{}14\textbar{}\allowbreak{}/\textbar{}\allowbreak{}12\textbar{}\allowbreak{}/\textbar{}\allowbreak{}2559
\item \tht{วิธีการอันชาญฉลาด} $\rightarrow$ \tht{วิธี}\textbar{}\allowbreak{}\tht{การ}\textbar{}\allowbreak{}\tht{อัน}\textbar{}\allowbreak{}\tht{ชาญ}\textbar{}\allowbreak{}\tht{ฉลาด}
\item \tht{การแปรรูปผลิตผลจากพืช} $\rightarrow$ \tht{การ}\textbar{}\allowbreak{}\tht{แปร}\textbar{}\allowbreak{}\tht{รูป}\textbar{}\allowbreak{}\tht{ผลิตผล}\textbar{}\allowbreak{}\tht{จาก}\textbar{}\allowbreak{}\tht{พืช}
\item \tht{โรงงานผลิตผลไม้ฟรีซดราย} $\rightarrow$ \tht{โรง}\textbar{}\allowbreak{}\tht{งาน}\textbar{}\allowbreak{}\tht{ผลิต}\textbar{}\allowbreak{}\tht{ผล}\textbar{}\allowbreak{}\tht{ไม้}\textbar{}\allowbreak{}\tht{ฟรีซดราย}
\item \tht{การสัมมนาเชิงวิชาการ} $\rightarrow$ \tht{การ}\textbar{}\allowbreak{}\tht{สัมมนา}\textbar{}\allowbreak{}\tht{เชิง}\textbar{}\allowbreak{}\tht{วิชาการ}
\item \tht{วิชาการท่องเที่ยวและการโรงแรม} $\rightarrow$ \tht{วิชา}\textbar{}\allowbreak{}\tht{การ}\textbar{}\allowbreak{}\tht{ท่อง}\textbar{}\allowbreak{}\tht{เที่ยว}\textbar{}\allowbreak{}\tht{และ}\textbar{}\allowbreak{}\tht{การ}\textbar{}\allowbreak{}\tht{โรง}\textbar{}\allowbreak{}\tht{แรม}
\item \tht{ประเดี๋ยวกระผมนำมาให้ขอรับ} $\rightarrow$ \tht{ประเดี๋ยว}\textbar{}\allowbreak{}\tht{กระผม}\textbar{}\allowbreak{}\tht{นำ}\textbar{}\allowbreak{}\tht{มา}\textbar{}\allowbreak{}\tht{ให้}\textbar{}\allowbreak{}\tht{ขอ}\textbar{}\allowbreak{}\tht{รับ}
\item \tht{เรียนเชิญคุณสมบัติ} \tht{เมทะนี} $\rightarrow$ \tht{เรียน}\textbar{}\allowbreak{}\tht{เชิญ}\textbar{}\allowbreak{}\tht{คุณ}\textbar{}\allowbreak{}\tht{สมบัติ}\textbar{}\allowbreak{} \textbar{}\allowbreak{}\tht{เมทะนี}
\item \tht{๑}) \tht{ผู้สมัครต้องมีคุณสมบัติดังนี้} $\rightarrow$ \tht{๑})\textbar{}\allowbreak{} \textbar{}\allowbreak{}\tht{ผู้}\textbar{}\allowbreak{}\tht{สมัคร}\textbar{}\allowbreak{}\tht{ต้อง}\textbar{}\allowbreak{}\tht{มี}\textbar{}\allowbreak{}\tht{คุณสมบัติ}\textbar{}\allowbreak{}\tht{ดัง}\textbar{}\allowbreak{}\tht{นี้}
\item \tht{อัตราเร็ว} \tht{๑}.\tht{๒}-\tht{๑}.\tht{๕} \tht{เมตร}/\tht{วินาที} $\rightarrow$ \tht{อัตรา}\textbar{}\allowbreak{}\tht{เร็ว}\textbar{}\allowbreak{} \textbar{}\allowbreak{}\tht{๑}.\tht{๒}\textbar{}\allowbreak{}-\textbar{}\allowbreak{}\tht{๑}.\tht{๕}\textbar{}\allowbreak{} \textbar{}\allowbreak{}\tht{เมตร}\textbar{}\allowbreak{}/\textbar{}\allowbreak{}\tht{วินาที}
\item “\tht{เอ}... \tht{ใช่ไหมนะ}? \tht{ใช่สิ}!” \tht{ดิฉันสับสน} $\rightarrow$ “\textbar{}\allowbreak{}\tht{เอ}\textbar{}\allowbreak{}.\textbar{}\allowbreak{}.\textbar{}\allowbreak{}.\textbar{}\allowbreak{} \textbar{}\allowbreak{}\tht{ใช่}\textbar{}\allowbreak{}\tht{ไหม}\textbar{}\allowbreak{}\tht{นะ}\textbar{}\allowbreak{}?\textbar{}\allowbreak{} \textbar{}\allowbreak{}\tht{ใช่}\textbar{}\allowbreak{}\tht{สิ}\textbar{}\allowbreak{}!\textbar{}\allowbreak{}”\textbar{}\allowbreak{} \textbar{}\allowbreak{}\tht{ดิฉัน}\textbar{}\allowbreak{}\tht{สับ}\textbar{}\allowbreak{}\tht{สน}
\item \tht{เขา} [\tht{อดีตนายกฯ}] \tht{ขออภัยโทษ} $\rightarrow$ \tht{เขา}\textbar{}\allowbreak{} \textbar{}\allowbreak{}[\textbar{}\allowbreak{}\tht{อดีต}\textbar{}\allowbreak{}\tht{นายก}\textbar{}\allowbreak{}\tht{ฯ}\textbar{}\allowbreak{}]\textbar{}\allowbreak{} \textbar{}\allowbreak{}\tht{ขอ}\textbar{}\allowbreak{}\tht{อภัยโทษ}
\item \tht{บ้านขนาด} > 70 \tht{เมตร}\^{}2 \tht{แต่} < 150 \tht{เมตร}\^{}2 \tht{ราคา} THB500,000 + \tht{ภาษี} 7\% $\rightarrow$ \tht{บ้าน}\textbar{}\allowbreak{}\tht{ขนาด}\textbar{}\allowbreak{} \textbar{}\allowbreak{}>\textbar{}\allowbreak{} \textbar{}\allowbreak{}70\textbar{}\allowbreak{} \textbar{}\allowbreak{}\tht{เมตร}\textbar{}\allowbreak{}\^{}\textbar{}\allowbreak{}2\textbar{}\allowbreak{} \textbar{}\allowbreak{}\tht{แต่}\textbar{}\allowbreak{} \textbar{}\allowbreak{}<\textbar{}\allowbreak{} \textbar{}\allowbreak{}150\textbar{}\allowbreak{} \textbar{}\allowbreak{}\tht{เมตร}\textbar{}\allowbreak{}\^{}\textbar{}\allowbreak{}2\textbar{}\allowbreak{} \textbar{}\allowbreak{}\tht{ราคา}\textbar{}\allowbreak{} \textbar{}\allowbreak{}THB\textbar{}\allowbreak{}500,000\textbar{}\allowbreak{} \textbar{}\allowbreak{}+\textbar{}\allowbreak{} \textbar{}\allowbreak{}\tht{ภาษี}\textbar{}\allowbreak{} \textbar{}\allowbreak{}7\textbar{}\allowbreak{}\%
\item \tht{บริษัท} Yahoo! \tht{มีมูลค่า} \$300,000-400,000 $\rightarrow$ \tht{บริษัท}\textbar{}\allowbreak{} \textbar{}\allowbreak{}Yahoo\textbar{}\allowbreak{}!\textbar{}\allowbreak{} \textbar{}\allowbreak{}\tht{มี}\textbar{}\allowbreak{}\tht{มูล}\textbar{}\allowbreak{}\tht{ค่า}\textbar{}\allowbreak{} \textbar{}\allowbreak{}\$\textbar{}\allowbreak{}300,000\textbar{}\allowbreak{}-\textbar{}\allowbreak{}400,000
\item \tht{ฝากซื้อแป้งสาลี}, \tht{ไข่ไก่}, \tht{เนย}, \tht{น้ำตาล} \tht{ฯลฯ} $\rightarrow$ \tht{ฝาก}\textbar{}\allowbreak{}\tht{ซื้อ}\textbar{}\allowbreak{}\tht{แป้ง}\textbar{}\allowbreak{}\tht{สาลี}\textbar{}\allowbreak{},\textbar{}\allowbreak{} \textbar{}\allowbreak{}\tht{ไข่}\textbar{}\allowbreak{}\tht{ไก่}\textbar{}\allowbreak{},\textbar{}\allowbreak{} \textbar{}\allowbreak{}\tht{เนย}\textbar{}\allowbreak{},\textbar{}\allowbreak{} \textbar{}\allowbreak{}\tht{น้ำ}\textbar{}\allowbreak{}\tht{ตาล}\textbar{}\allowbreak{} \textbar{}\allowbreak{}\tht{ฯลฯ}
\item \tht{ติด} \#\tht{ผัดไทเส้นจันทน์} \tht{แล้วแอดไลน์} @lineID $\rightarrow$ \tht{ติด}\textbar{}\allowbreak{} \textbar{}\allowbreak{}\#\textbar{}\allowbreak{}\tht{ผัด}\textbar{}\allowbreak{}\tht{ไท}\textbar{}\allowbreak{}\tht{เส้น}\textbar{}\allowbreak{}\tht{จันทน์}\textbar{}\allowbreak{} \textbar{}\allowbreak{}\tht{แล้ว}\textbar{}\allowbreak{}\tht{แอด}\textbar{}\allowbreak{}\tht{ไลน์}\textbar{}\allowbreak{} \textbar{}\allowbreak{}@lineID
\item {[(18*67)+15]}-17\^{}2 = \~{}42.22 $\rightarrow$ [\textbar{}\allowbreak{}(\textbar{}\allowbreak{}18\textbar{}\allowbreak{}*\textbar{}\allowbreak{}67\textbar{}\allowbreak{})\textbar{}\allowbreak{}+\textbar{}\allowbreak{}15\textbar{}\allowbreak{}]\textbar{}\allowbreak{}-\textbar{}\allowbreak{}17\textbar{}\allowbreak{}\^{}\textbar{}\allowbreak{}2\textbar{}\allowbreak{} \textbar{}\allowbreak{}=\textbar{}\allowbreak{} \textbar{}\allowbreak{}\~{}\textbar{}\allowbreak{}42.22
\item \tht{ประชาชนต่อสู้เพื่อประชาธิปไตย} $\rightarrow$ \tht{ประชา}\textbar{}\allowbreak{}\tht{ชน}\textbar{}\allowbreak{}\tht{ต่อ}\textbar{}\allowbreak{}\tht{สู้}\textbar{}\allowbreak{}\tht{เพื่อ}\textbar{}\allowbreak{}\tht{ประชาธิปไตย}
\item \tht{พิธีพระราชทานปริญญาบัตร} \tht{จุฬาลงกรณ์มหาวิทยาลัย} $\rightarrow$ \tht{พิธี}\textbar{}\allowbreak{}\tht{พระ}\textbar{}\allowbreak{}\tht{ราชทาน}\textbar{}\allowbreak{}\tht{ปริญญาบัตร}\textbar{}\allowbreak{} \textbar{}\allowbreak{}\tht{จุฬาลงกรณ์มหาวิทยาลัย}
\item \tht{เด็กกำพร้าในบ้านโกโรโกโส} $\rightarrow$ \tht{เด็ก}\textbar{}\allowbreak{}\tht{กำพร้า}\textbar{}\allowbreak{}\tht{ใน}\textbar{}\allowbreak{}\tht{บ้าน}\textbar{}\allowbreak{}\tht{โกโรโกโส}
\item \tht{กำพร้าไว้แน่น} \tht{ๆ} \tht{อย่าให้หล่น} $\rightarrow$ \tht{กำ}\textbar{}\allowbreak{}\tht{พร้า}\textbar{}\allowbreak{}\tht{ไว้}\textbar{}\allowbreak{}\tht{แน่น}\textbar{}\allowbreak{} \textbar{}\allowbreak{}\tht{ๆ}\textbar{}\allowbreak{} \textbar{}\allowbreak{}\tht{อย่า}\textbar{}\allowbreak{}\tht{ให้}\textbar{}\allowbreak{}\tht{หล่น}
\item \tht{สุภาพสตรีหัวเราะคิกคัก} $\rightarrow$ \tht{สุภาพสตรี}\textbar{}\allowbreak{}\tht{หัว}\textbar{}\allowbreak{}\tht{เราะ}\textbar{}\allowbreak{}\tht{คิกคัก}
\item \tht{รำคาญเสียงคิกคิกคักคัก} $\rightarrow$ \tht{รำคาญ}\textbar{}\allowbreak{}\tht{เสียง}\textbar{}\allowbreak{}\tht{คิก}\textbar{}\allowbreak{}\tht{คิก}\textbar{}\allowbreak{}\tht{คัก}\textbar{}\allowbreak{}\tht{คัก}
\item 2542. \tht{สิทธิชุมชนและอัตลักษณ์ทางวัฒนธรรม}: \tht{ชาติพันธุ์วรรณาในปทุมธานี} [\tht{เอกสารสัมภาษณ์ประกอบเวทีเสวนา}]. \tht{ชุมชนในสายธารโลกาภิวัตน์}. $\rightarrow$ 2542\textbar{}\allowbreak{}.\textbar{}\allowbreak{} \textbar{}\allowbreak{}\tht{สิทธิ}\textbar{}\allowbreak{}\tht{ชุม}\textbar{}\allowbreak{}\tht{ชน}\textbar{}\allowbreak{}\tht{และ}\textbar{}\allowbreak{}\tht{อัตลักษณ์}\textbar{}\allowbreak{}\tht{ทาง}\textbar{}\allowbreak{}\tht{วัฒนธรรม}\textbar{}\allowbreak{}:\textbar{}\allowbreak{} \textbar{}\allowbreak{}\tht{ชาติพันธุ์วรรณา}\textbar{}\allowbreak{}\tht{ใน}\textbar{}\allowbreak{}\tht{ปทุมธานี}\textbar{}\allowbreak{} \textbar{}\allowbreak{}[\textbar{}\allowbreak{}\tht{เอกสาร}\textbar{}\allowbreak{}\tht{สัมภาษณ์}\textbar{}\allowbreak{}\tht{ประกอบ}\textbar{}\allowbreak{}\tht{เวที}\textbar{}\allowbreak{}\tht{เสวนา}\textbar{}\allowbreak{}]\textbar{}\allowbreak{}.\textbar{}\allowbreak{} \textbar{}\allowbreak{}\tht{ชุม}\textbar{}\allowbreak{}\tht{ชน}\textbar{}\allowbreak{}\tht{ใน}\textbar{}\allowbreak{}\tht{สาย}\textbar{}\allowbreak{}\tht{ธาร}\textbar{}\allowbreak{}\tht{โลกาภิวัตน์}\textbar{}\allowbreak{}.
\item \tht{นกโป่งวิดจะจับคู่ทำรังในช่วงฤดูฝน} $\rightarrow$ \tht{นก}\textbar{}\allowbreak{}\tht{โป่งวิด}\textbar{}\allowbreak{}\tht{จะ}\textbar{}\allowbreak{}\tht{จับ}\textbar{}\allowbreak{}\tht{คู่}\textbar{}\allowbreak{}\tht{ทำ}\textbar{}\allowbreak{}\tht{รัง}\textbar{}\allowbreak{}\tht{ใน}\textbar{}\allowbreak{}\tht{ช่วง}\textbar{}\allowbreak{}\tht{ฤดู}\textbar{}\allowbreak{}\tht{ฝน}
\item \tht{นกกาเหว่ามักจะแอบวางไข่ในรังของอีกา} $\rightarrow$ \tht{นก}\textbar{}\allowbreak{}\tht{กาเหว่า}\textbar{}\allowbreak{}\tht{มัก}\textbar{}\allowbreak{}\tht{จะ}\textbar{}\allowbreak{}\tht{แอบ}\textbar{}\allowbreak{}\tht{วาง}\textbar{}\allowbreak{}\tht{ไข่}\textbar{}\allowbreak{}\tht{ใน}\textbar{}\allowbreak{}\tht{รัง}\textbar{}\allowbreak{}\tht{ของ}\textbar{}\allowbreak{}\tht{อี}\textbar{}\allowbreak{}\tht{กา}
\item \tht{ถิ่นอาศัยของนกเค้าแมวมีหลากหลายคล้ายกับนกฮูก} $\rightarrow$ \tht{ถิ่น}\textbar{}\allowbreak{}\tht{อาศัย}\textbar{}\allowbreak{}\tht{ของ}\textbar{}\allowbreak{}\tht{นก}\textbar{}\allowbreak{}\tht{เค้า}\textbar{}\allowbreak{}\tht{แมว}\textbar{}\allowbreak{}\tht{มี}\textbar{}\allowbreak{}\tht{หลาก}\textbar{}\allowbreak{}\tht{หลาย}\textbar{}\allowbreak{}\tht{คล้าย}\textbar{}\allowbreak{}\tht{กับ}\textbar{}\allowbreak{}\tht{นก}\textbar{}\allowbreak{}\tht{ฮูก}
\item \tht{พนักงานขนส่งสินค้าขับรถรับ}-\tht{ส่งพัสดุ} $\rightarrow$ \tht{พนัก}\textbar{}\allowbreak{}\tht{งาน}\textbar{}\allowbreak{}\tht{ขน}\textbar{}\allowbreak{}\tht{ส่ง}\textbar{}\allowbreak{}\tht{สิน}\textbar{}\allowbreak{}\tht{ค้า}\textbar{}\allowbreak{}\tht{ขับ}\textbar{}\allowbreak{}\tht{รถ}\textbar{}\allowbreak{}\tht{รับ}\textbar{}\allowbreak{}-\textbar{}\allowbreak{}\tht{ส่ง}\textbar{}\allowbreak{}\tht{พัสดุ}
\item \tht{กระทรวงมหาดไทยมีหน้าที่บริหารราชการแผ่นดินส่วนกลาง} $\rightarrow$ \tht{กระทรวง}\textbar{}\allowbreak{}\tht{มหาด}\textbar{}\allowbreak{}\tht{ไทย}\textbar{}\allowbreak{}\tht{มี}\textbar{}\allowbreak{}\tht{หน้า}\textbar{}\allowbreak{}\tht{ที่}\textbar{}\allowbreak{}\tht{บริหาร}\textbar{}\allowbreak{}\tht{ราชการ}\textbar{}\allowbreak{}\tht{แผ่น}\textbar{}\allowbreak{}\tht{ดิน}\textbar{}\allowbreak{}\tht{ส่วน}\textbar{}\allowbreak{}\tht{กลาง}
\item \tht{นายจุติเสฏฐ์ไม่มีเบาะแสของทองคำแท่ง} $\rightarrow$ \tht{นาย}\textbar{}\allowbreak{}\tht{จุติ}\textbar{}\allowbreak{}\tht{เสฏฐ์}\textbar{}\allowbreak{}\tht{ไม่}\textbar{}\allowbreak{}\tht{มี}\textbar{}\allowbreak{}\tht{เบาะ}\textbar{}\allowbreak{}\tht{แส}\textbar{}\allowbreak{}\tht{ของ}\textbar{}\allowbreak{}\tht{ทอง}\textbar{}\allowbreak{}\tht{คำ}\textbar{}\allowbreak{}\tht{แท่ง}
\item \tht{บ้านขนาด} > 70 \tht{เมตร}\^{}2 \tht{ราคา} THB500,000 + \tht{ภาษี} 7\% $\rightarrow$ \tht{บ้าน}\textbar{}\allowbreak{}\tht{ขนาด}\textbar{}\allowbreak{} \textbar{}\allowbreak{}>\textbar{}\allowbreak{} \textbar{}\allowbreak{}70\textbar{}\allowbreak{} \textbar{}\allowbreak{}\tht{เมตร}\textbar{}\allowbreak{}\^{}\textbar{}\allowbreak{}2\textbar{}\allowbreak{} \textbar{}\allowbreak{}\tht{ราคา}\textbar{}\allowbreak{} \textbar{}\allowbreak{}THB\textbar{}\allowbreak{}500,000\textbar{}\allowbreak{} \textbar{}\allowbreak{}+\textbar{}\allowbreak{} \textbar{}\allowbreak{}\tht{ภาษี}\textbar{}\allowbreak{} \textbar{}\allowbreak{}7\textbar{}\allowbreak{}\%
\item \tht{ถนนเลียบคลองสอง} $\rightarrow$ \tht{ถนน}\textbar{}\allowbreak{}\tht{เลียบ}\textbar{}\allowbreak{}\tht{คลอง}\textbar{}\allowbreak{}\tht{สอง}
\item \tht{นายภาคภูมิเดินทางมา} $\rightarrow$ \tht{นาย}\textbar{}\allowbreak{}\tht{ภาค}\textbar{}\allowbreak{}\tht{ภูมิ}\textbar{}\allowbreak{}\tht{เดิน}\textbar{}\allowbreak{}\tht{ทาง}\textbar{}\allowbreak{}\tht{มา}
\item \tht{ทำงานขะมักเขม้น} $\rightarrow$ \tht{ทำ}\textbar{}\allowbreak{}\tht{งาน}\textbar{}\allowbreak{}\tht{ขะมัก}\textbar{}\allowbreak{}\tht{เขม้น}
\item \tht{อนึ่ง} \tht{ขอแจ้งว่า} $\rightarrow$ \tht{อนึ่ง}\textbar{}\allowbreak{} \textbar{}\allowbreak{}\tht{ขอ}\textbar{}\allowbreak{}\tht{แจ้ง}\textbar{}\allowbreak{}\tht{ว่า}
\end{itemize}
}

\section{Downstream Results in Full}
\label{app:downstream}
\begin{table*}[!htb]
\centering
\caption{\textbf{Downstream results in full.} BiLSTM taggers on LST20, scored by entity F1 for NER and token
accuracy for POS, and a BiLSTM classifier on Wongnai sentiment, scored by accuracy and macro-F1 on the
official test set, all trained without pretraining on the output of each segmenter; mean $\pm$ standard deviation over 10 seeds. $\Delta$ is khammoon
minus the segmenter in the row on the same seeds, with the $p$-value of a paired Wilcoxon signed-rank test.}
\label{tab:downstream}
\small
\setlength{\tabcolsep}{4pt}
\resizebox{\textwidth}{!}{%
\begin{tabular}{l rr rr rr rr}
\toprule
 & \multicolumn{4}{c}{\textbf{LST20}} & \multicolumn{4}{c}{\textbf{Wongnai}} \\
\cmidrule(lr){2-5}\cmidrule(lr){6-9}
\textbf{Segmenter} & \textbf{NER F1} $\uparrow$ & $\Delta$ & \textbf{POS acc.} $\uparrow$ & $\Delta$ & \textbf{Accuracy} $\uparrow$ & $\Delta$ & \textbf{Macro-F1} $\uparrow$ & $\Delta$ \\
\midrule
characters & 0.675 $\pm$ 0.003 & +0.044 ($p<0.01$) & 0.952 $\pm$ 0.001 & +0.004 ($p<0.01$) & 0.349 $\pm$ 0.031 & +0.084 ($p<0.01$) & 0.276 $\pm$ 0.022 & +0.091 ($p<0.01$) \\
PyThaiNLP newmm & \textbf{0.727} $\pm$ 0.004 & -0.008 ($p=0.01$) & \textbf{0.956} $\pm$ 0.001 & +0.001 ($p=0.28$) & 0.423 $\pm$ 0.020 & +0.011 ($p=0.32$) & 0.363 $\pm$ 0.012 & +0.004 ($p=0.28$) \\
DeepCut & 0.666 $\pm$ 0.012 & +0.054 ($p<0.01$) & 0.953 $\pm$ 0.005 & +0.003 ($p=0.04$) & 0.425 $\pm$ 0.028 & +0.009 ($p=0.38$) & 0.365 $\pm$ 0.025 & +0.002 ($p=0.85$) \\
AttaCut & 0.658 $\pm$ 0.011 & +0.061 ($p<0.01$) & 0.953 $\pm$ 0.003 & +0.003 ($p<0.01$) & 0.424 $\pm$ 0.028 & +0.010 ($p=0.62$) & 0.366 $\pm$ 0.026 & +0.001 ($p=1.00$) \\
\textbf{khammoon} & 0.719 $\pm$ 0.007 & -- & \textbf{0.956} $\pm$ 0.002 & -- & \textbf{0.434} $\pm$ 0.020 & -- & \textbf{0.367} $\pm$ 0.015 & -- \\
\bottomrule
\end{tabular}}
\end{table*}

\Cref{tab:downstream} lists the numbers behind \Cref{fig:downstream}. $\Delta$ is khammoon minus
the segmenter in the row on the same ten seeds, and $p$ is from a paired Wilcoxon signed-rank
test over those seeds.

\section{Inter-Annotator Agreement and Adjudication}
\label{app:iaa}
\paragraph{Agreement.} \Cref{tab:iaa} gives the agreement of the three annotators, A, B and C, on the
400 sentences per corpus that all three annotated independently. Fleiss' kappa treats every position
between two characters as an item with two categories, boundary or no boundary; pairwise F1 is the
boundary F1 of one annotator against another. The last three columns count sentences by whether all
three, exactly two, or none of the annotators produced the same segmentation.

\begin{table}[!tb]
\centering
\caption{\textbf{Inter-annotator agreement} on 400 triple-annotated sentences per corpus.}
\label{tab:iaa}
\small
\setlength{\tabcolsep}{4pt}
\begin{tabular}{l r rrr rrr}
\toprule
 & & \multicolumn{3}{c}{\textbf{Pairwise F1}} & \multicolumn{3}{c}{\textbf{Sentences (\%)}} \\
\cmidrule(lr){3-5}\cmidrule(lr){6-8}
\textbf{Corpus} & \textbf{Fleiss' $\kappa$} & A--B & A--C & B--C & all 3 & 2 of 3 & none \\
\midrule
BEST2009 & 0.986 & 0.993 & 0.988 & 0.989 & 60.5 & 25.8 & 13.8 \\
Wangchan & 0.991 & 0.995 & 0.994 & 0.992 & 45.5 & 38.5 & 16.0 \\
\bottomrule
\end{tabular}
\end{table}

\paragraph{Where annotators disagreed.} The disputed strings fall into five groups, listed with
the most frequent examples.
\begin{itemize}[nosep]
  \item \textbf{Pali--Sanskrit compounds whose parts also occur as Thai words:}
        \tht{เหตุการณ์}, \tht{ประสบการณ์}, \tht{วัตถุประสงค์}, \tht{อุดมการณ์}, \tht{วิชาการ},
        \tht{องค์การ}, \tht{ชาติพันธุ์}, \tht{คุณภาพ}, \tht{ปัญญาประดิษฐ์}.
  \item \textbf{Everyday compounds and couplets near the keep/split line:} \tht{หัวเราะ},
        \tht{หน้าที่}, \tht{ได้แก่}, \tht{เอ็นดู}, \tht{ต้องการ}, \tht{อันดับ}, \tht{อ่อนโยน},
        \tht{ร่วมกับ}, \tht{ลูกค้า}, \tht{วงจร}.
  \item \textbf{Proper names, place names and transliterated brand names:} \tht{สยามแม็คโคร},
        \tht{โรบินสันไลฟ์สไตล์}, \tht{มิตรทาวน์}, \tht{ไอคอนสยาม}, \tht{ไทยพาณิชย์},
        \tht{ขอนแก่น}, \tht{ประสิทธิ์ศักดิ์}.
  \item \textbf{Abbreviations, numbering and symbols:} \tht{ผบช.ภ.}, \tht{กป.อพช.},
        \tht{รอง ผบ.ตร.}, \texttt{COVID-19}, telephone numbers, \tht{(1)}, \texttt{50\%}.
  \item \textbf{Sound-pattern words:} \tht{กะปริบกะปรอย}, \tht{อากัปกริยา}.
\end{itemize}

\paragraph{Adjudication and consistency.} The reference data were finalized in four steps.
\begin{enumerate}[nosep]
  \item The three versions of each triple-annotated sentence were reconciled into one, and the
        remaining 1,852 BEST sentences and 1,600 Wangchan documents were divided among the three
        annotators.
  \item The linguists reviewed every sentence and edited 1,293 of the 4,252, 186 of them BEST and
        1,107 Wangchan.
  \item A consistency check over all 4,252 sentences listed every pair of adjacent tokens
        $X$\textbar{}$Y$ that also occurs as one token $XY$ somewhere in the data, 663 pairs in all.
        The 19 pairs in which the minority form occurs at most twice were put to the linguist: 8
        were rewritten to the majority form everywhere, for example \tht{ถูก|ต้อง} with 453
        occurrences against one merged \tht{ถูกต้อง}, \tht{ล่า|สุด}, \tht{โซเชียลมีเดีย} and
        \tht{ราชอาณาจักร}, and 11 were declared context-dependent or left open and resolved
        sentence by sentence, among them \tht{ทำลาย} with 51 merged against 1 split, \tht{วิชาการ},
        and heading numbers such as \tht{2)} against \tht{2|)}. Six further pairs with a nearly
        even count, such as \tht{พัฒนาการ} with 26 merged and 7 split, were also resolved sentence
        by sentence. After this step 13 pairs
        still occur in both forms, all of them deliberately.
  \item A final check confirmed that every sentence concatenates back to its source text exactly,
        with no mismatches, that the whitespace of the source is preserved, and that the data contain
        13,951 distinct token types, 7,518 of them Thai-only.
\end{enumerate}
The annotation outputs, the agreement script and the disagreement lists are part of the released
data.

\end{document}